\documentclass[10pt,twocolumn,letterpaper]{article}

\usepackage{cvpr}              % To produce the CAMERA-READY version

\usepackage{amsfonts} % Blackboard Math
\usepackage{amsmath} % DeclareMathOperator

\usepackage{adjustbox}
\usepackage{algorithm}
\usepackage{algpseudocode}

\usepackage{mathtools}

\DeclareMathOperator*{\argmin}{argmin} % thin space, limits underneath in displays

\usepackage{amsthm}
\usepackage{thmtools}
\usepackage{thm-restate} % restate therorem
\newtheorem{definition}{Definition}

\usepackage{booktabs}       % professional-quality tables
\usepackage{multicol}
\usepackage{multirow}
\usepackage{makecell}       % newline within cell

\newcommand{\bbE}{\mathbb{E}} % Expectation
\definecolor{cvprblue}{rgb}{0.21,0.49,0.74}
\usepackage[accsupp]{axessibility}
\usepackage[pagebackref,breaklinks,colorlinks,allcolors=cvprblue]{hyperref}
\usepackage{arydshln}

\newcommand{\gt}[1]{\textcolor{gray}{#1}}

\def\paperID{3950} % *** Enter the Paper ID here
\def\confName{CVPR}
\def\confYear{2025}

\title{Sufficient Invariant Learning for Distribution Shift}

\author{Taero Kim\textsuperscript{1}, Subeen Park\textsuperscript{1}, Sungjun Lim\textsuperscript{1}, Yonghan Jung\textsuperscript{2}, Krikamol Muandet\textsuperscript{3}, Kyungwoo Song\textsuperscript{1}\thanks{Corresponding Author}\\
\textsuperscript{1}Yonsei University, \textsuperscript{2}Purdue University, \textsuperscript{3}CISPA Helmholtz Center for Information Security\\
{\tt\small taero.kim@yonsei.ac.kr, sallyna602@yonsei.ac.kr, lsj9862@yonsei.ac.kr, } \\[-1mm]{\tt\small jung222@purdue.edu, muandet@cispa.de, kyungwoo.song@yonsei.ac.kr}}

\begin{document}
\maketitle

\begin{abstract}
    Learning robust models under distribution shifts between training and test datasets is a fundamental challenge in machine learning. While learning invariant features across environments is a popular approach, it often assumes that these features are fully observed in both training and test sets—a condition frequently violated in practice. When models rely on invariant features absent in the test set, their robustness in new environments can deteriorate. To tackle this problem, we introduce a novel learning principle called the Sufficient Invariant Learning (SIL) framework, which focuses on learning a sufficient subset of invariant features rather than relying on a single feature. After demonstrating the limitation of existing invariant learning methods, we propose a new algorithm, Adaptive Sharpness-aware Group Distributionally Robust Optimization (ASGDRO), to learn diverse invariant features by seeking common flat minima across the environments. We theoretically demonstrate that finding a common flat minima enables robust predictions based on diverse invariant features. Empirical evaluations on multiple datasets, including our new benchmark, confirm ASGDRO's robustness against distribution shifts, highlighting the limitations of existing methods. Code: ~\url{https://github.com/MLAI-Yonsei/SIL-ASGDRO}.

\end{abstract}

\section{Introduction} \label{sec1}
Machine learning models typically assume that training and test data are drawn from the same distribution. However, in real-world scenarios, this assumption is often violated whenever the training and test distribution differ, known as distribution shifts. In these cases, model performance tends to degrade, highlighting the need to develop models that are robust to distribution shifts for reliable outcomes.

\begin{figure}[t]
    \centering
    \includegraphics[width=0.45\textwidth]{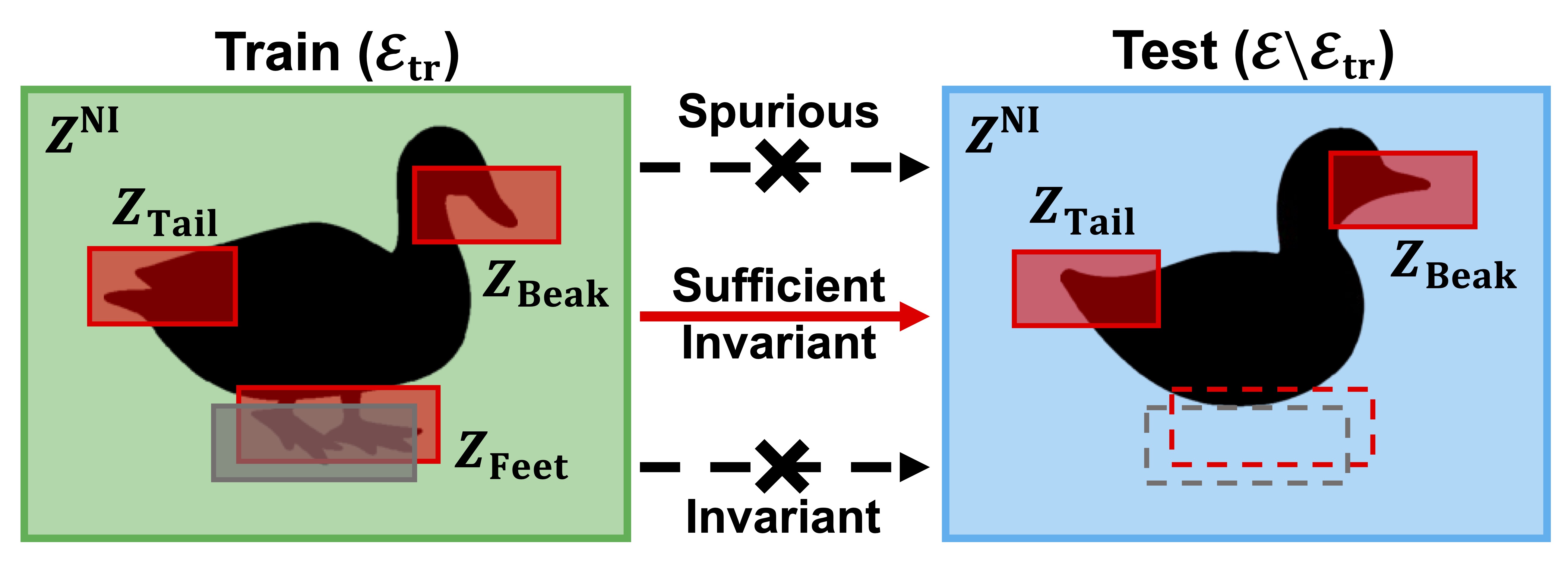}
    \vspace{-1.0em}
    \caption{\small Left visualizes the images that contain a spurious feature, $Z^{\text{NI}}$, and multiple invariant features, $Z_{\text{Tail}}$, $Z_{\text{Beak}}$, and $Z_{\text{Feet}}$ in training environment $\mathcal{E}_{\text{tr}}$. If the model focuses on the $Z^{\text{NI}}$ (green background), then it fails to predict correctly in the test environment $\mathcal{E}\backslash \mathcal{E}_{\text{tr}}$ (Right). Even if the model captures the invariant features in $\mathcal{E}_{\text{tr}}$, e.g., $Z_{\text{Feet}}$, it still fails to predict correctly when the invariant features are not present (Gray). However, it is possible to predict correctly if we learn diverse invariant features sufficiently, $Z_{\text{Feet}}$, $Z_{\text{Tail}}$, and $Z_{\text{Beak}}$. With SIL (Red), the model predicts the label using remaining invariant features, $Z_{\text{Tail}}$ and $Z_{\text{Beak}}$ even though $Z_{\text{Feet}}$ is not present in the test environment $\mathcal{E}\backslash \mathcal{E}_{\text{tr}}$.}
    \label{fig:fig1} 
    \vspace{-2em}
\end{figure}

To train models robust to distribution shift, invariant learning focuses on identifying latent features that remain constant across environments, referred to as invariant features. These features enable consistent predictions across environments by discouraging models from relying on spurious features \cite{arjovsky2019invariant} -- features that are not preserved across changes in environments or groups\footnote{In this paper, the terms \textit{environment} and \textit{domain} are used interchangeably. A \textit{group} refers to a subpopulation corresponding to a particular label within a specific environment.}. 
For example, in domain generalization tasks \cite{koh2021wilds,gulrajani2020search}, the goal is to learn invariant features that consistently predict labels across multiple environments. Assuming that the learned invariant features persist in all unseen environments, they guarantee the model's generalization performance on new environments \cite{muandet2013domain,li2018deep}. 
Similarly, learning models robust to subpopulation shifts is essential in cases of severe imbalances between groups. In this scenario, invariant features play a crucial role in addressing the challenges faced by underrepresented groups, which are disproportionately impacted by strong spurious correlations \citep{sagawa2019distributionally,yao2022improving,izmailov2018averaging}.

However, learning all possible invariant features is challenging in practice because most existing invariant learning approaches focus on eliminating spurious correlations, which can be achieved by leveraging only a subset of the invariant features present in the training environments. Moreover, invariant features identified by the model may not be observable in unseen environments \cite{guo2024outofvariable,tsymbal2004problem}. This underscores the importance of learning a \textit{sufficient} number of invariant features, rather than relying on a single invariant feature. To address this, we introduce a novel approach called \textit{Sufficient Invariant Learning} (SIL), which focuses on learning a sufficient set of invariant features for improved generalization. 
For example, consider the scenario depicted in \cref{fig:fig1}. Training environments for an image of a bird may include multiple invariant features, such as $Z_{\text{Tail}}$, $Z_{\text{Beak}}$ and $Z_{\text{Feet}}$. If a model relies on a single invariant feature, say $Z_{\text{Feet}}$, it may fail to classify an image of the bird if the feature is unobservable (e.g., the bird’s feet are hidden underwater). In contrast, if the model uses a sufficiently diverse set of invariant features (e.g.,  all of $Z_{\text{Tail}}$, $Z_{\text{Beak}}$ and $Z_{\text{Feet}}$), it can still classify the image correctly as long as one or more of the other invariant features are present. This highlights the robustness and generalization benefits of learning a sufficient number of invariant features.

In this study, we develop the SIL framework and demonstrate that leveraging sufficiently diverse invariant features through SIL enhances model robustness. As a method for SIL, we propose \textit{Adaptive Sharpness-aware Group Distributionally Robust Optimization} (ASGDRO). We show that ASGDRO attains SIL by effectively learning diverse invariant features while successfully eliminating spurious correlations. Furthermore, we show that the ability of ASGDRO to perform SIL is due to its convergence to a common flat minima \cite{foret2020sharpness}  across diverse environments. Through empirical evaluations on a toy example and our newly introduced SIL benchmark dataset, we show that existing invariant learning algorithms fall short in capturing diverse invariant features, whereas ASGDRO successfully achieves SIL. By learning a wide range of invariant features sufficiently, ASGDRO exhibits robust generalization performance under various distribution shift scenarios, as evidenced by extensive experiments involving subpopulation and domain shifts.

\section{Related Works} \label{sec2}
\subsection{Invariant Learning for Distribution Shift}
The standard approach to modern deep learning is Empirical Risk Minimization (ERM) \cite{vapnik99}, which minimizes the average training loss. However, ERM may not guarantee robustness in distribution shifts. To improve the generalization performance in distribution shift, Group Distributionally Robust Optimization (GDRO) minimizes the worst group loss for each iteration to alleviate spurious correlations \cite{sagawa2019distributionally}. Meanwhile, various studies utilize loss gradient for invariant learning. For example, \citet{arjovsky2019invariant} minimizes the gradient norm of the fixed classifier across environments. Other research matches the loss gradient for each environment to find invariant features \cite{shi2021gradient,rame2022fishr}. 
Furthermore, balancing the representation using selective sampling with mix-up samples \cite{yao2022improving} or re-training the classifier on a small balanced set \cite{kirichenko2022last} show the effectiveness of learning a robust model. Some studies enhance generalization by combining invariant learning algorithms with feature extractors with rich representations \cite{zhang2022rich,chen2024understanding,zhang2023learning} or resolving the conflict between ERM and invariant learning objectives \cite{chen2022pareto}.

Under the assumption that invariant features in the training environment also exist in the test environment, invariant learning theoretically guarantees an optimal predictor \cite{rojas2018invariant}. However, we argue that existing invariant learning algorithms do not learn sufficiently diverse invariant features, and they still suffer significant performance drops in test environments where some invariant features are unobserved \cite{guo2024outofvariable,tsymbal2004problem}. \citet{lin2023spurious} consider settings with multiple features; however, they solely address scenarios where only spurious features are multiple in nature. To remedy this problem, we introduce the novel framework, SIL, and guarantee the generalization ability for diverse invariant features. Through experiments on the newly proposed benchmark in this paper, as well as on existing benchmarks for distribution shifts \cite{gulrajani2020search,koh2021wilds}, we demonstrate that our novel algorithm designed for SIL leads to more robust predictions.

\subsection{Flatness and Generalization}
Various studies argue that finding flat minima improves generalization performance \cite{keskar2016large,neyshabur2017exploring}. As a result, many algorithms emerge to find flat minima. Sharpness-aware Minimization (SAM) \cite{foret2020sharpness} finds flat minima by minimizing the maximum training loss of neighborhoods for the current parameter within $\rho$ radius ball on the parameter space. Moreover, Adaptive SAM (ASAM) introduces the normalization operator to get a better correlation between flatness and the model's generalization ability by avoiding the scale symmetries between the layers \cite{kwon2021asam}. Stochastic Weight Averaging (SWA) also reaches the flat minima by averaging the weight \cite{izmailov2018averaging}. Under the IID setting, these approaches \cite{foret2020sharpness,kwon2021asam,izmailov2018averaging} successfully decrease the generalization gap.

\citet{cha2021swad} shows that optimizing the model towards flatter minima through weight averaging improves domain generalization ability. However, it is still necessary to verify whether the models operate robustly through weight averaging when strong spurious correlations exist. Indeed, some studies demonstrate that weight averaging may still not be robust in certain subpopulation shift tasks \cite{rame2022diverse}. \citet{zhang2023flatness} also shows that flat minima make the models more robust to the noise. However, our study focuses on the effectiveness of flatness in more extreme distribution shift settings, such as subpopulation shift and domain generalization. \citet{springer2024sharpnessaware} presents that when easy-to-learn and hard-to-learn features coexist, models trained by SAM learn balanced representations. This aligns with our observations, and we aim to achieve SIL by removing spurious correlations and learning sufficiently diverse invariant features by introducing the constraints related to flatness.

\section{Methodology} \label{sec3}
\subsection{Problem Setting}
Let $\mathcal{X}$, $\mathcal{Y}$, $\mathcal{Z}$, and $\Theta$ denote the input, label, feature, and parameter spaces, respectively. Consider a set of environments $\mathcal{E}$, where each environment $e \in \mathcal{E}$ is associated with a dataset $\mathcal{D}^{e} =\{(X_{i}^{e},Y_{i}^{e})\}_{i=1}^{n_{e}}$, with $X_{i}^{e}\in\mathcal{X}$, $Y_{i}^{e}\in\mathcal{Y}$, and $n_{e}$ indicating the number of data points in $e$. We assume a feature set \(Z = (Z^{\text{I}}, Z^{\text{NI}}) \subset \mathcal{Z}\), where \(Z^{\text{I}}\) denotes invariant features that satisfy the following invariance condition, and \(Z^{\text{NI}}\) denotes spurious features whose correlation with \(Y^e\) varies across environments \(e\)~\cite{arjovsky2019invariant,creager2021environment,krueger2021out}.
\begin{definition}[Invariance Condition] \label{def:invariance}
$Z^{\mathrm{I}}$ is a set of invariant features satisfying
\begin{equation}
    \bbE[Y^e|Z^{\mathrm{I}}]=\bbE[Y^{e'}|Z^{\mathrm{I}}] \quad \text{for all $e, e' \in \mathcal{E}_\mathrm{tr}$}, \label{eq:invariance} \nonumber
\end{equation}
$\mathcal{E}_{\mathrm{tr}}\subset \mathcal{E}$ denotes the set of training environments.
\end{definition}
\noindent We denote an invariant feature $Z^{\text{I}}_i$ as the singleton set containing $i$th element of $Z^{\text{I}}$, where $i \in \{1, \dots, p\}$ and $p$ is the number of invariant features. In \cref{fig:fig1}, the invariant features are $Z^{\text{I}}=\{Z_{\text{Beak}}, Z_{\text{Tail}}, Z_{\text{Feet}}\}$ with $p=3$, the spurious feature is $Z^{\text{NI}}=\{Z_{\text{Background}}\}$ and one example of an invariant feature is $Z^{\text{I}}_{1}=\{Z_{\text{Feet}}\}$, corresponding to the feet.

Suppose a model $f = h \circ g $ parametrized by $\theta = (\theta_{g}, \theta_{h})\in \Theta$, where $g: \mathcal{X} \rightarrow \mathcal{Z}$ is an encoder with parameters $\theta_{g}$ and $h: \mathcal{Z} \rightarrow \mathcal{Y}$ is a classifier with parameters  $\theta_{h}$. Let  $\mathcal{R}^{e}(\theta)=\mathbb{E}[\ell (f(X^{e}; \theta), Y^{e})]$ denote the risk of a model $f$ in environment $e$, where $\ell$ denotes a loss function. Invariant learning seeks to minimize the maximum risk across environments,
\begin{equation}\label{eq:inv_loss}
\min_{\theta}\max_{e \in \mathcal{E}} \mathcal{R}^{e}(\theta),
\end{equation}
% or, in terms of $Z$, $\mathcal{R}^{\text{I}}(\theta; Z):=\max_{e \in \mathcal{E}} \mathcal{R}^{e}(\theta;Z)$.
and to train models that have robust performance and generalization ability for unseen environments by learning invariant features $Z^{\text{I}}$ \cite{arjovsky2019invariant,sagawa2019distributionally,creager2021environment,krueger2021out}. In particular, given $Z^{\text{I}}$, \citet{rojas2018invariant} demonstrate that learning optimal classifier $\theta_h^*$, which is based on all invariant features in $Z^{\text{I}}$, leads to robust model predictions, i.e.,
\begin{equation} \label{eq:il}
    \theta_h^* \in \min_{\theta_h}\max_{e\in\mathcal{E}} \mathcal{R}^{e}(\theta_h),
    \end{equation}
where $\mathcal{R}^{e}(\theta_h)=\mathbb[\ell(h(Z^{\text{I}};\theta_{h}),Y^{e})]$, assuming that the invariance condition holds for all $e \in \mathcal{E}$.

%the optimal classifier for \cref{eq:il} is not unique

\subsection{Sufficient Invariant Learning} \label{sec:sil}
While models trained via invariant learning have shown effectiveness under various distribution shifts, this does not imply that the optimal classifier satisfying \cref{eq:il} is unique. For $\mathcal{E}_{\text{tr}}$, \cref{def:invariance} holds for any subset $\hat{Z}^{\text{I}}\subseteq Z^{\text{I}}$ and any classifier relying on $\hat{Z}^{\text{I}}$ can be optimal.
% To address the challenges in such scenarios, we propose a framework called Sufficient Invariant Learning (SIL). 
% In the classification task, we concentrate on the fact that any subset of $Z^{\text{I}}$ allows a model to minimize the maximum risk across environments $e \in \mathcal{E}_{\text{tr}}$. 
In \cref{fig:fig1}, the classifier may utilize only \( Z_{\text{Feet}} \), or it may employ all \( Z^{\text{I}}_{i} \) simultaneously in \( \mathcal{E}_{\text{tr}} \), to distinguish between waterbirds from landbirds. Therefore, the optimal encoder minimizing \cref{eq:inv_loss} is also not unique, as it depends on the non-unique optimal classifier \cite{arjovsky2019invariant}. To distinguish predictive mechanisms using different $\hat{Z}^{\text{I}}$, we define the invariant mechanism.
\begin{definition}[Invariant Mechanism] \label{def:invmec}
For an encoder $g_{\theta_{g}^{\mathrm{I}}}$ parameterized by $\theta_{g}^{\mathrm{I}}$ and a classifier $h_{\theta_{h}^{\mathrm{I}}}$ parameterized by $\theta_{h}^{\mathrm{I}}$, 
the invariant mechanism $\theta^{\mathrm{I}} = (\theta^{\mathrm{I}}_{g}, \theta^{\mathrm{I}}_{h}) \in \Theta$  is a tuple for a subset $\hat{Z}^{\mathrm{I}}\subseteq Z^{\mathrm{I}}$ satisfying the followings: 
\begin{align}
    \text{Condition 1: }& h_{\theta_{h}^{\mathrm{I}}}:\hat{Z}^{\mathrm{I}}\mapsto Y^{e}, \quad \forall e \in \mathcal{E}_{\mathrm{tr}}. \nonumber  \\
    \text{Condition 2: }& \theta^{\mathrm{I}}\in \argmin_{\theta}\max_{e \in \mathcal{E}_{\mathrm{tr}}} \mathcal{R}^{e}(\theta).
    \nonumber
\end{align}
\end{definition}
\noindent Specifically, we denote the invariant mechanism that utilizes only $Z^{\text{I}}_{i}$ as $\theta^{\text{I}}_i$, for $i=\{1, \dots p\}$. Invariant mechanisms that rely solely on a specific invariant feature $\theta_i^{\text{I}}$ may struggle to make robust predictions when the part of the input corresponding to that feature is corrupted by noise, missing due to cropping, or occluded by environmental factors. This non-uniqueness suggests that training encoders via classifier invariance \cite{arjovsky2019invariant,ahuja2021invariance} or enhancing them to capture richer information \cite{zhang2022rich,chen2024understanding} can benefit from additional regularization to leverage other invariant features. This observation also implies that robust optimization methods designed to minimize \cref{eq:inv_loss} over $\mathcal{E}_\text{tr}$ \cite{duchi2016statistics,oren2019distributionally,sagawa2019distributionally} have an avenue for achieving enhanced generalization performance.

We argue that training more robust models requires ensuring generalization across sufficiently diverse sets of invariant features. To this end, we introduce a novel invariant learning framework, termed Sufficient Invariant Learning (SIL), which encourages learning diverse invariant features:
\begin{definition}[Sufficient Invariant Learning] \label{def:sic}
Sufficient Invariant Learning refers to identify $\theta^{\mathrm{SI}}$ such that 
\begin{align} 
\theta^{\mathrm{SI}} &\in \argmin_{\theta}\max_{e\in\mathcal{E}}\mathcal{R}^{e}(\theta), \nonumber \\ 
s.t. \quad \theta^{\mathrm{SI}}_{h} &\in \argmin_{\theta_h} \max_{e \in \mathcal{E}} \max_{\hat{Z}^{\mathrm{I}} \subseteq Z^{\mathrm{I}}} \mathbb{E}[\ell(h_{\theta_h}(\hat{Z}^{\mathrm{I}}),Y^{e})]. \nonumber 
\label{eq:sil}
\end{align}
\end{definition}

\noindent SIL aims to train a classifier that performs robustly not only across all environments but also with respect to any subset $\hat{Z}^{\text{I}}$. It encourages the model to leverage sufficiently diverse invariant features, assuming that representations of these features have already been learned from the target task \cite{kirichenko2022last}. The main challenge in achieving SIL lies in the cost of obtaining individually intervened data for each $\hat{Z}^{\text{I}}$. To achieve this, we propose ASGDRO, a novel method inspired by the geometry of the loss surface, which promotes SIL by identifying common flat minima.

\subsection{ASGDRO: Adaptive Sharpness-aware Group Distributionally Robust Optimization} \label{sec:sec3.3}

In the literature on model merging and multi-task learning \cite{ilharco2022editing,wortsman2022robust,ainsworth2022git,rame2023model}, it is often assumed that a robust model across all tasks lies within the linear interpolation of models that perform well on each individual task. Inspired by this observation, we consider $\theta^{\text{I}}_i$ as a model that performs well on a single task, and we hypothesize that $\theta^{\text{SI}}$ exists within the linear interpolation of these mechanisms. Without loss of generality, subsets that are not singletons can be equivalently represented as an interpolation of singleton invariant features $Z^{\text{I}}_i$. Hence, for the remainder of this work, we restrict our consideration to $Z^{\text{I}}_i$ and $Z^{\text{I}}$ (Appendix \ref{append:wlog}). The key difference from previous studies is that we evaluate each task solely on the same dataset. Therefore, as discussed in \cref{sec:sil}, different invariant mechanisms are expected to have similar risks, 
 \begin{equation}
     \mathcal{R}^{e}(\theta^{\text{SI}})-\mathcal{R}^{e}(\theta^{\text{I}}_{i})\approx 0 \quad \text{for all $e\in \mathcal{E}_{\text{tr}}$}. \nonumber 
 \end{equation}

 \noindent A challenge for SIL is that we do not have access to information about $\theta^{\text{I}}_i$. However, based on the observation in \citet{neyshabur2020being} that different models trained from the same pre-trained model lie in the same loss basin, we assume that models located on the linear path between \( \theta^{\text{SI}} \) and $\theta^{\text{I}}_i$ also exhibit similar risk. Therefore, $\theta^{\text{SI}}$ should guarantee low risks within a ball of radius at least $\max_{i} || \theta^{\text{I}}_i - \theta^{\text{SI}} ||$, denoted as $\rho$, in Euclidean space. Introducing a perturbation $\epsilon_e := \theta^{\text{I}}_i - \theta^{\text{SI}}$, we obtain the following condition for the risk of $\theta^{\text{I}}_i$:
\begin{equation}
\max_{i \in \{1, \dots, p\}} \mathcal{R}^{e}(\theta^{\text{I}}_i) = \max_{||\epsilon_e|| \leq \rho}{\mathcal{R}^{e}(\theta^{\text{SI}}+\epsilon_e}).     \nonumber 
\end{equation}
From our motivation, $\rho$ is a hyper-parameter adjusting the model class of $\theta^{\text{I}}_{i}$ deviated from $\theta^{\text{SI}}$. Moreover, according to \cref{def:invariance}, all $\theta^{\text{I}}_i$ should exhibit robust performance across environments $e \in \mathcal{E}_{\text{tr}}$. Finally, we propose a novel objective function named Adaptive Sharpness-aware Group Distributionally Robust Optimization (ASGDRO), which is formulated as follows:
    \begin{equation} \label{eq:asgdro}
        \max_{e \in \mathcal{E}_{tr}}\max_{||\epsilon_{e}|| \leq \rho} \mathcal{R}^{e}(\theta + \epsilon_{e}).
    \end{equation}
In the following sections, we theoretically show that ASGDRO not only learns invariant features but also balances the learning of invariant mechanisms, thereby achieving SIL. Also, we demonstrate that ASGDRO finds the common flat minima across environments, leading to SIL.

\begin{figure*}[t!]
\centering
    \includegraphics[width=\textwidth]{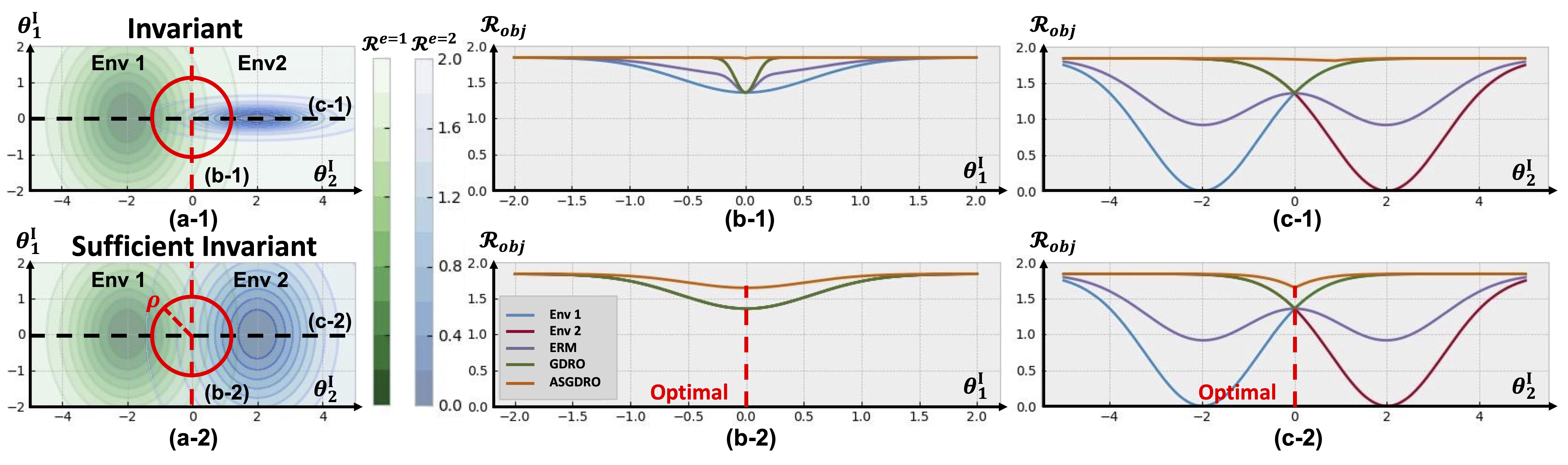}
    \vspace{-2em}
    \caption{\small \textbf{Sufficient Invariant Learning and Common Flat Minima} In (a-1) and (a-2), two axes, $\theta_1^{\text{I}}$ and $\theta_2^{\text{I}}$, represent the invariant directions of parameters corresponding to each invariant mechanism respectively. The red circle indicates the area bound by $\rho$ for measuring flatness in ASGDRO. (b-1) and (b-2) show that when Env 2 has sharp minima in the direction of $\theta_1^{\text{I}}$, GDRO still converges, but ASGDRO does not have any optimal point due to the sharpness of  $\theta_1^{\text{I}}$. However, in (c-1) and (c-2) when both invariant directions of Env 2 as well as Env 1 are flat, ASGDRO has an optimal point and prefers to converge. That is, ASGDRO learns diverse invariant features sufficiently.}
    \label{fig:fig2}
    \vspace{-1em}
\end{figure*}

\begin{algorithm}[t]
    \caption{ASGDRO}\label{alg:ASGDRO}
    \small
    \begin{algorithmic}[1]
    \Require Training dataset $D_{\text{tr}}^{e}=\{(X^{e},Y^{e})\}$ for  $e\in\mathcal{E}_\text{tr}$, Radius $\rho>0$, Learning rate $\eta>0$, Robust step size $\gamma>0$, The number of environments $|\mathcal{E}_{\text{tr}}|$, Normalization Matrix $T_{\theta}$.
    \State Initialization: $\theta_{0}$; $\lambda_{e}^{(0)}=1/|\mathcal{E}_{\text{tr}}|$;
    \For {$t=1,2,3,\ldots$}
    \For {$e=1,\ldots, \lvert\mathcal{E}_{\text{tr}} \rvert $}
    \State Compute training loss $\mathcal{R}^{e}(\theta_{t})$;
    \State Compute $\epsilon_{e}^{*} = \rho\frac{T_{\theta}^{2}\nabla\mathcal{R}^{e}(\theta_{t})}{\left \| T_{\theta}\nabla\mathcal{R}^{e}(\theta_{t}) \right \|}$;
    \State Gradient ascent: ${\theta_{t}^{*}}$ = ${\theta_{t}}+\epsilon_{e}^{*}$;
    \State Find loss for each environment $\mathcal{R}^{e}({\theta_{t}^{*}})$;
    \State Compute $\tilde{\lambda}_{e}^{(t)}=\lambda_{e}^{(t-1)}\exp(\gamma \mathcal{R}^{e}({\theta_{t}^{*}}))$; 
    \State Return to ${\theta_{t}}$;
    \EndFor
    \State Update $\lambda_{e}^{(t)}={\tilde{\lambda}_{e}^{(t)}}/{\sum_{e}{\tilde{\lambda}_{e}^{(t)}}}$;
    \State Compute $\mathcal{R}_{\text{ASGDRO}}({\theta_{t}})=\sum_{e}{\lambda_{e}^{(t)}}\mathcal{R}^{e}({\theta_{t}^{*}})$;
    \State Compute $\nabla{\mathcal{R}_{\text{ASGDRO}}({\theta_{t}})}=\sum_{e}{\lambda_{e}^{(t)}}\nabla \mathcal{R}^{e}({\theta_{t}^{*}})$;
    \State Return to ${\theta_{t}}$;
    \State Update the parameters: $\theta_{t+1} = {\theta_{t}} - \eta \nabla \mathcal{R}_{\text{ASGDRO}}({\theta_{t}})$;
    \EndFor
    \end{algorithmic}
\end{algorithm}

 \subsection{SIL and Common Flat Minima}
We demonstrate that ASGDRO trains the model to achieve SIL by showing that ASGDRO balances the use of diverse invariant mechanisms.
\begin{restatable}{theorem}{thm}
\label{thm}
Let $\theta^{\mathrm{I}}_\lambda$ be a convex combination of $\theta^{\mathrm{I}}_{i}$, where $\lambda$ is a $p$-dimensional vector.  
Consider mean-squared error as the loss function. 
Assume a linear model with $Z \in \mathbb{R}^p$, where the $p$ features are orthogonal, and suppose $Z=Z^{\mathrm{I}}=(1, \dots, 1)$. Then, 
\begin{align}
\lambda^{*}= &\argmin_{\lambda}\max_{e \in \mathcal{E}_{\text{tr}}} \max_{||\epsilon|| \leq \rho} \mathcal{R}^{e}(\theta^{\mathrm{I}}_\lambda + \epsilon) \nonumber  \\
\approx &\argmin_{\lambda}\max_{e \in \mathcal{E}_{\text{tr}}} \left[ \mathcal{R}^{e}(\theta^{\mathrm{I}}_\lambda)+\rho||\lambda|| \cdot||\nabla\mathcal{R}^{e}(\theta^{\mathrm{I}}_\lambda)|| \right] \label{eq:reg} \\
= &\argmin_{\lambda} ||\lambda|| = (\frac{1}{p}, \ldots, \frac{1}{p})  \nonumber 
\end{align}
where $||\cdot||$ denotes $L_2$ norm.
\end{restatable}

\noindent Refer to Appendix \ref{append:proof_theorem} for the proof. \cref{thm} states that ASGDRO ensures that even when invariant features contribute equally to the output, the model does not favor a simple solution focusing on a single invariant feature. Instead, it learns a diverse range of invariant mechanisms. As shown in \cref{eq:reg}, this regularization effect arises through the gradient norm $||\nabla\mathcal{R}^{e}(\theta)||$.
\begin{restatable}{proposition}{propsecond} 
% \label{prop2}
By the Taylor expansion, 
    \begin{equation} \label{commonflat}
        \max_{e \in \mathcal{E}}\max_{||\epsilon_{e}|| \leq \rho} \mathcal{R}^{e}(\theta + \epsilon_{e}) \approx \max_{e\in \mathcal{E}}[\mathcal{R}^{e}(\theta)+ \rho|| \nabla\mathcal{R}^{e}(\theta) ||]. \nonumber 
    \end{equation}
ASGDRO leads to a regularization of the gradient norm, $\mathcal{R}^{e}$, $|| \nabla\mathcal{R}^{e}(\theta) ||$, across environments, which drives the model to converge to common flat minima.
\end{restatable}
\noindent Refer to Appendix \ref{append:proof_prop} for proof. As demonstrated in \citep{zhao2022penalizing}, small $|| \nabla\mathcal{R}^{e}(\theta)||$ indicates flat minima. We also demonstrate this property empirically in \cref{fig:hess_celeba} and Appendix \ref{append:hess}. Finally, we argue that finding common flat minima encourages the model to learn sufficiently diverse invariant mechanisms. Moreover, this aligns with existing studies in IID settings, which suggest that flatter minima improve the generalization performance of models \cite{foret2020sharpness,kwon2021asam,keskar2016large}. Additionally, we demonstrate in Appendix \ref{append:spurious} that ASGDRO successfully eliminates the spurious feature $Z^{e}$ while effectively learning the invariant feature.

\subsection{Implementation of ASGDRO}
From \citet{foret2020sharpness}, maximum value of inner term in \cref{eq:asgdro} is approximated when $\epsilon_{e}= \rho\frac{\nabla\mathcal{R}^{e}(\theta)}{\left \| \nabla\mathcal{R}^{e}(\theta) \right \|}$. However, \citet{kwon2021asam} show that by introducing the normalization matrix $T_{\theta}$, which removes the scale symmetry present on the loss surface, the correlation between flatness and generalization performance is strengthened. ASGDRO also adopts the same $T_{\theta}$, and modified objective function is as follows:
\begin{equation}
    \mathcal{R}_{\text{ASGDRO}}(\theta)=\max_{e\in\mathcal{E}_{\text{tr}}}\mathcal{R}^{e}(\theta+\epsilon_{e}^*),\quad \nonumber 
\end{equation}
\noindent where $\epsilon_{e}^*=\rho\frac{T_{\theta}^{2}\nabla\mathcal{R}^{e}(\theta)}{\left \| T_{\theta}\nabla\mathcal{R}^{e}(\theta) \right \|}$ is an adversarial perturbation for each environment $e$, $T_{\theta}=\operatorname{diag}(\operatorname{concat}(\|\bf{k}_1\|\mathbf{1}_{\bf{n}(\bf{k}_1)}, $ $\dots,\|\bf{k}_m\|\mathbf{1}_{\bf{n}(\bf{k}_m)},\lvert\omega_{1}\rvert, \dots, \lvert \omega_{q}\rvert))$, where $\bf{k}_m$ denotes a convolution kernel, $\omega_{\bf{q}}$ represents other parameters and $\bf{n}(\cdot)$ indicates the number of parameters. 

To address the instability in training that arises from the optimization approach of selecting only the worst environment at each step, we adopt an alternative gradient-based optimization algorithm inspired by GDRO \cite{sagawa2019distributionally}. We modify ASGDRO into the form of linear interpolation across environments and update their coefficients: 
{\small \begin{equation} 
\max_{e\in\mathcal{E}_\text{tr}}{\mathcal{R}^e(\theta + \epsilon_{e}^{*})} = \max_{\sum_{e}{\lambda_{e}}=1,\lambda_e\geq0}{\sum_{e\in\mathcal{E}_\text{tr}}\lambda_{e} \mathcal{R}^e(\theta + \epsilon_{e}^{*}),} \nonumber
 \end{equation}}
\noindent where $\lambda_{e}$ is the weight imposed on adversarial perturbed loss for each environment. Finally, we update our model parameter from the current parameter $\theta_{t}$ as follows:
{\small \begin{equation}
    \theta_{t} - \eta \nabla\mathcal{R}_{\text{ASGDRO}}(\theta) = \theta_{t} - \eta \sum_{e\in\mathcal{E}_\text{tr}}\lambda_{e}^{(t)}\nabla\mathcal{R}^{e}(\theta_{t}+\epsilon_{e}^{*}),\nonumber
\end{equation}}
\noindent where $\eta$ denotes the learning rate and $\lambda_{e}^{(t)}$ denotes the weight imposed on each loss of environment at time step $t$. Refer to \cref{alg:ASGDRO} for the details. In practice, for computational efficiency, in all experiments except for the toy example, instead of calculating $\epsilon_{e}^* = \rho\frac{T_{\theta}^{2}\nabla\mathcal{R}^{e}(\theta)}{\left | T_{\theta}\nabla\mathcal{R}^{e}(\theta) \right |}$ for each environment, we use a common adversarial perturbation utilizing the empirical risk $\mathcal{R}_{S}(\theta) = \frac{1}{\lvert D^{e} \rvert \lvert \mathcal{E}_{\text{tr}}\rvert} \sum_{e\in\mathcal{E}_{\text{tr}}}\sum_{n_{e}} \ell (f(X^{e};\theta),Y^{e})$, i.e. $\epsilon^{*}=\rho\frac{T_{\theta}^{2}\nabla\mathcal{R}_{S}(\theta)}{\left | T_{\theta}\nabla\mathcal{R}_{S}(\theta) \right |}$. As a result, the gradient ascending through $\epsilon^{*}$ is performed only once regardless of the number of environments, and the loss for each environment is evaluated using the same perturbed parameters, $\theta + \epsilon^{*}$.

\section{Experiments} \label{sec4}
\begin{figure*}[!ht]
\centering
    \includegraphics[width=\linewidth]{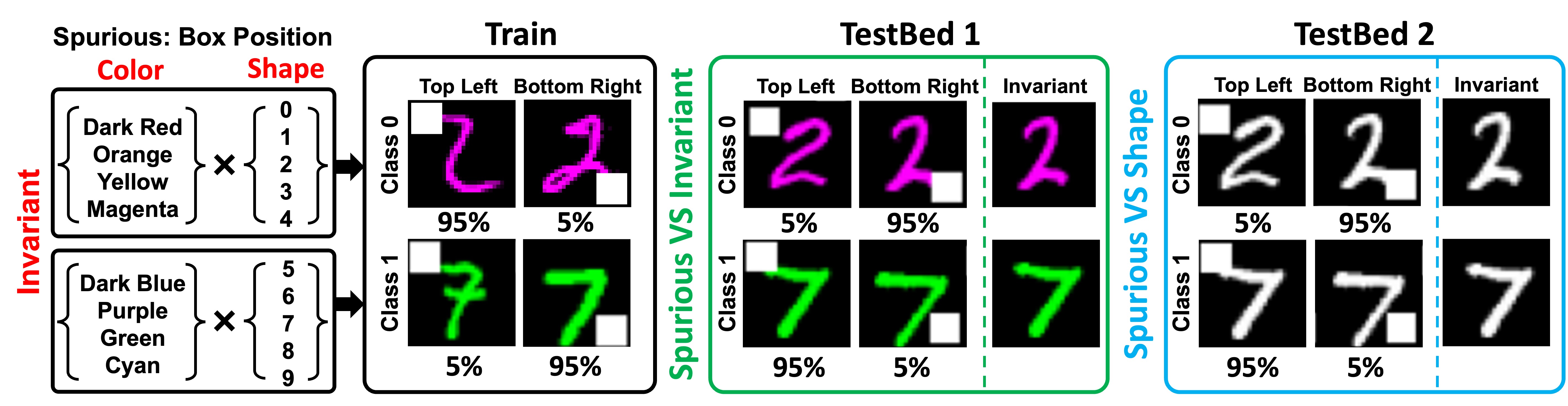}
    \caption{\small \textbf{Overview of H-CMNIST.} There are three features, color and shape (invariant features, $Z^{\text{I}}=\{Z_{\text{color}}, Z_{\text{shape}}\}$) and box position (spurious feature, $Z^{\text{NI}}=\{Z_{\text{BP}}\}$). The ratio of $Z_{\text{BP}}$ is flipped between the train and test set. The test set consists of two testbeds, one for evaluating whether learning invariant features and the other for evaluating whether learning sufficiently diverse invariant features.}
    \label{fig:h-cmnist_overview}
\end{figure*}
\subsection{Toy Exmaple} \label{sec:sec3.2}
We demonstrate through a toy example that the representative invariant learning algorithm GDRO \cite{sagawa2019distributionally} fails to learn diverse invariant mechanisms, whereas ASGDRO successfully achieves SIL by encouraging the model to converge to the common flat minima (\cref{fig:fig2}). First, we assume that we know two different directions corresponding to the different invariant mechanism $\theta_{1}^{\text{I}}$ and $\theta_{2}^{\text{I}}$, which learns different invariant features, $Z^{\text{I}}_1$ and $Z^{\text{I}}_2$, respectively. We define the loss surface of each environment $e$ following a Gaussian function with respect to $\theta_{1}^{\text{I}}$ and $\theta_{2}^{\text{I}}$:
{\small
\begin{align}
G(\theta) = \frac{1}{2\pi\sqrt{\left|\Sigma\right|}} \exp\left(-\frac{1}{2}(\theta - \mu)^T \Sigma^{-1} (\theta - \mu)\right), \nonumber \\
\text{where} \quad \theta  = \begin{bmatrix} \theta_{1}^{\text{I}} \\ \theta_{2}^{\text{I}} \end{bmatrix}, \mu^{(e)} = \begin{bmatrix} \mu_1 \\ \mu_2 \end{bmatrix}, \Sigma^{(e)} = \begin{bmatrix} \sigma_{11} \sigma_{12} \\ \sigma_{21}  \sigma_{22} \end{bmatrix}. \nonumber
\end{align}}
To make losses greater than 0, we subtracted $G(\theta)$ from its maximum value. As a result, we define the loss surface corresponding to the two environments, each with a minimum value of 0, as follows:
{\small
\begin{align}
    \mathcal{R}^{e=1}(\theta) = \max_{\theta}\left[G(\theta; \mu^{(1)}, \Sigma^{(1)})\right] - G(\theta; \mu^{(1)}, \Sigma^{(1)}) \nonumber \\ 
    \mathcal{R}^{e=2}(\theta) = \max_{\theta}\left[G(\theta; \mu^{(2)}, \Sigma^{(2)})\right] - G(\theta; \mu^{(2)}, \Sigma^{(2)})  \nonumber
\end{align}}

\begin{table} 
\begin{adjustbox}{width=\linewidth}
        \begin{tabular}{cccp{0.05cm}cc}
        \toprule
        &\multicolumn{2}{c}{TestBed 1} & & \multicolumn{2}{c}{TestBed 2} \\ \cmidrule{2-3} \cmidrule{5-6}
       & Spu \& Inv                   & Inv                     &  & Spu \& Shape                      & Shape                          \\ \midrule
        ERM    &  97.11   $\pm$ 3.44 & 98.75   $\pm$ 1.19 & & 34.64  $\pm$ 9.90  & 57.41  $\pm$ 2.58  \\
        ASAM   &  98.57   $\pm$ 1.21 & 98.12   $\pm$ 1.74 & & 34.78   $\pm$ 8.41 & 57.07   $\pm$ 1.91 \\
        GDRO   &  \textbf{99.95  $\pm$ 0.07}  & \textbf{99.92  $\pm$ 0.08}  & & \underline{57.53  $\pm$ 2.11} & \underline{61.44   $\pm$ 1.03}  \\
        ASGDRO & \underline{99.88   $\pm$ 0.11} & \underline{99.83   $\pm$ 0.12} & & \textbf{66.62   $\pm$ 5.61} & \textbf{69.17   $\pm$ 6.19} \\ \bottomrule
        \end{tabular}
        \end{adjustbox}
        \vspace{-0.4em}
        \caption{\small \textbf{H-CMNIST Results.} TestBed 1 evaluates whether the model learns easy invariant feature $Z_{\text{color}}$, and TestBed 2 evaluates the ability to learn additional invariant feature $Z_{\text{shape}}$.}
\label{tab:h-cmnist_result}
        \vspace{-1.5em}
\end{table}

Now we create sharp or flat minima in a specific direction by adjusting the covariance matrix $\Sigma^{(e)}$. In this example, we consider a fixed situation where both $e=1$ and $e=2$ have flat minima with respect to $\theta_{2}^{\text{I}}$. When $\mathcal{R}^{e=1}(\theta)$ always has flat minima in the direction of $\theta_{1}^{\text{I}}$, we aim to observe how the loss $\mathcal{R}_{obj}$ corresponding to each objective function changes depending on whether $\theta_{2}^{\text{I}}$ has sharp or flat minima (a-1 and a-2 in \cref{fig:fig2}). 
The parameters that we use to generate the toy examples are as follows:
{\small
\begin{align}
    \text{Env 1 ($e=1$) :} \hspace{0.5em}  \mu = \begin{bmatrix} -2.0 \\ 0.0 \end{bmatrix},& \hspace{0.5em} \text{Env 2 ($e=2$) :} \hspace{0.5em} \mu = \begin{bmatrix} 2.0 \\ 0.0 \end{bmatrix}\nonumber \\
    \text{Flat :} \hspace{0.5em}  \Sigma  = \begin{bmatrix} 1.5 & 0.0 \\ 0.0 & 2.0 \end{bmatrix},& \hspace{0.5em} \text{Sharp :} \hspace{0.5em}   \Sigma  = \begin{bmatrix} 1.5 & 0.0 \\ 0.0 & 0.05 \end{bmatrix} \nonumber
\end{align}
}

We evaluate each algorithm through the loss surface in each direction (second and third columns of \cref{fig:fig2}). When Env 2 exhibits sharpness for $\theta_{1}^{\text{I}}$ (first row of \cref{fig:fig2}), it indicates that learning the invariant feature corresponding to $\theta_{1}^{\text{I}}$ may result in a large generalization gap \citep{keskar2016large}. However, GDRO does not incorporate regularization on flatness and only considers the loss at the current parameter, allowing convergence to a sharp solution. From \cref{thm}, it implies the large gradient norm, and this situation does not constitute successful SIL. In contrast, ASGDRO, which takes into account the loss in neighboring parameters, avoids sharp regions for $\theta_{1}^{\text{I}}$ (b-1 and c-1 in \cref{fig:fig2}).

When Env 2 is flat for $\theta_{1}^{\text{I}}$ (second row in \cref{fig:fig2}), we say that the model performs SIL if it converges into the common flat minima between Env 1 and Env 2. However, GDRO has the same loss at the optimal point in this situation as in the previous case, indicating that GDRO does not specifically regularize the model to perform SIL. On the other hand, ASGDRO, by accounting for common flat minima, identifies an optimal parameter that promotes learning of diverse invariant mechanisms (b-2 and c-2 in \cref{fig:fig2}). As a result, by considering flatness, the model performs SIL and is expected to make robust predictions in unseen environments by leveraging multiple invariant features.

\subsection{Heterogenous ColoredMNIST} \label{sec:sufficient}

By finding the common flat minima, ASGDRO learns diverse invariant features. To demonstrate this, we propose Heterogeneous ColoredMNIST (H-CMNIST), a new dataset designed to evaluate whether the model learns diverse invariant mechanisms sufficiently (\cref{fig:h-cmnist_overview}). H-CMNIST evaluate whether the remaining invariant feature is additionally learned by the algorithm, assuming that the model has already learned one invariant feature.

H-CMNIST includes two invariant features, the color $Z^{\text{I}}_1 =\{Z_{\text{color}}\}$ and shape of digits $Z^{\text{I}}_2 =\{Z_{\text{shape}}\}$, and one spurious feature, the position of the box (BP) $Z^{\text{NI}} =\{Z_{\text{BP}}\}$. That is, each class has its own colors and shapes.
Using BP, we construct two environments, Top Left (Env 0) and Bottom Right (Env 1). We design a scenario where spurious correlations occur \citep{sagawa2019distributionally,degrave2021ai}. Specifically, in the training set, 95\% of Left Top BP belongs to class 0, and only 5\% belongs to class 1. In contrast, we collect 95\% of Right Bottom BP in class 1, and assigned only 5\% to class 0.In the test sets, the composition of BP is flipped. That is, $Z_{\text{BP}}$ has a strong correlation with each class in the training set, but it does not hold in the test set. Refer to Appendix \ref{append:hcmnist} for details.

\cref{tab:h-cmnist_result} shows the results of H-CMNIST. H-CMNIST assumes an easily learnable invariant feature $Z_{\text{color}}$ to evaluate whether the model, having already learned one invariant feature, can learn additional invariant features $Z_{\text{shape}}$. Concretely, TestBed 1 serves as a preliminary step to verify that an easily learnable invariant feature is indeed present. In TestBed1, the performance of all algorithms is similar regardless of the presence of the spurious feature $Z_{\text{BP}}$, indicating that all have learned at least one invariant feature.

\begin{table}[t] 
  \centering
  \adjustbox{max width=\linewidth}{
  
    \begin{tabular}{lc@{ }cc@{ }cc@{ }cc@{ }c}
    \toprule
    \multicolumn{1}{c}{} &  \multicolumn{2}{c}{{CMNIST}} &\multicolumn{2}{c}{{Waterbirds}} & \multicolumn{2}{c}{{CelebA}} & \multicolumn{2}{c}{{CivilComments}}  \\ \cmidrule(lr){2-3} \cmidrule(lr){4-5} \cmidrule(lr){6-7} \cmidrule(lr){8-9}
    \multicolumn{1}{c}{}&\gt{Avg.} &Worst &\gt{Avg.} &Worst &\gt{Avg.} &Worst &Avg. &Worst  \\
    \midrule
     ERM$^\ddag$       &  \gt{27.8\%}  & 0.0\%                  & \gt{97.0\%} & 63.7\%               & \gt{94.9\%} & 47.8\%         & \gt{92.2\%} & 56.0\%   \\
     ASAM              &  \gt{40.5\%}  & 34.1\%                 & \gt{97.4\%} & 72.4\%               & \gt{93.7\%} & 46.5\%         & \gt{92.3\%} & 58.9\% \\
     IRM$^\ddag$       &   \gt{72.1\%} & 70.3\%                 & \gt{87.5\%} & 75.6\%               & \gt{94.0\%} & 77.8\%         & \gt{88.8\%} & 66.3\% \\
     IB-IRM$^\ddag$    &  \gt{72.2\%}  & 70.7\%                 & \gt{88.5\%} & 76.5\%               & \gt{93.6\%} & 85.0\%         & \gt{89.1\%} & 65.3\%  \\
     V-REx$^\ddag$     &  \gt{71.7\%}  & 70.2\%                 & \gt{88.0\%} & 73.6\%               & \gt{92.2\%} & 86.7\%         & \gt{90.2\%} & 64.9\%  \\
     CORAL$^\ddag$     &  \gt{71.8\%}  & 69.5\%                 & \gt{90.3\%} & 79.8\%               & \gt{93.8\%} & 76.9\%         & \gt{88.7\%} & 65.6\% \\
     GDRO$^\ddag$      &  \gt{72.3\%}  & 68.6\%                 & \gt{91.8\%} & {\underline{90.6\%}} & \gt{92.1\%} & 87.2\%         & \gt{89.9\%} & 70.0\%  \\
     DomainMix$^\ddag$ &  \gt{51.4\%}  & 48.0\%                 & \gt{76.4\%} & 53.0\%               & \gt{93.4\%} & 65.6\%         & \gt{90.9\%} & 63.6\% \\
     Fish$^\ddag$      &  \gt{46.9\%}  & 35.6\%                 & \gt{85.6\%} & 64.0\%               & \gt{93.1\%} & 61.2\%         & \gt{89.8\%} & 71.1\%  \\
     LISA$^\ddag$      &  \gt{74.0\%}  & {\underline{73.3\%}}   & \gt{91.8\%} & 89.2\%               & \gt{92.4\%} & \underline{89.3\%}         & \gt{89.2\%} & {\textbf{72.6\%}}  \\ 
    \midrule
     ASGDRO            &  \gt{74.8\%}  & {\textbf{74.2\%}}      & \gt{92.3\%} & \textbf{91.4\%}    & \gt{92.1\%} & \textbf{91.0\%} & \gt{90.2\%} & {\underline{71.8\%}}  \\ 
    \bottomrule
    \end{tabular}}
\caption{\small \textbf{Subpopulation Shift}. $\ddag$ denotes the performance reported from \citep{yao2022improving}. Avg. denotes average accuracy, and Worst denotes worst group accuracy. Refer to Appendix \ref{append:subpop} for details.} \label{tab:subpop}
\vspace{-2em}
\end{table}

However, in Testbed 2, without $Z_{\text{color}}$, both ERM and ASAM show significant performance discrepancies depending on the presence of spurious feature $Z_{\text{BP}}$. 
Compared with the results of TestBed 1, ERM and ASAM only learn $Z_{\text{color}}$ successfully, but they fail to capture the additional invariant feature, $Z_{\text{shape}}$. 
It indicates that even when a relatively easier invariant feature exists, the spurious feature influences the relatively more challenging invariant feature. Although GDRO exhibits robustness to spurious correlations compared to ERM and ASAM, it still fails to learn one of the invariant features, $Z_{\text{shape}}$. However, ASGDRO makes robust predictions against spurious features and more successful learning of shape features in TestBed2, compared to other baselines. It implies that SIL is necessary for the robust model and ASGDRO optimizes the model to learn sufficiently diverse invariant features $Z^{\text{I}}=\{Z_{\text{color}},Z_{\text{shape}}\}$ considering the common flat minima across environments.

\subsection{Experimental Results} \label{sec:subpopulation_shift}

In each result, boldface and underlined text denote the highest and second-highest accuracy for each dataset, respectively. Additional experiments, including efficiency or sensitivity analysis of ASGDRO can be found in the Appendix.

\begin{table}[t]
    \centering
    \begin{adjustbox}{width=\linewidth}
        \begin{tabular}{r@{}c@{}l@{ }cccccc}    
    \toprule
            \multirow{2}{*}{\textbf{PT}} &\multirow{2}{*}{\textbf{--}} & \multirow{2}{*}{\textbf{FT}}
                                                     & {\textbf{Camelyon17}}              & {\textbf{CivilComments}}           & {\textbf{FMoW}}                    & {\textbf{Amazon}}                     & {\textbf{RxRx1}}                          \\ 
            & &                                      & Avg. (\%)                          & Worst (\%)                         & Worst (\%)                         & 10th per. (\%)                        & Avg. (\%)                                 \\ 
    \midrule
             $\times$ &--& ERM                       & $70.3$ {\small$\pm${6.4}}          & $56.0$ {\small$\pm${3.6}}          & $32.3$ {\small$\pm${1.3}}          & \underline{$53.8$} {\small$\pm${0.8}} & $29.9$  {\small$\pm${0.4}}                \\
            $\times$ &--& GDRO                       & $68.4$ {\small$\pm${7.3}}          & $70.0$ {\small$\pm${2.0}}          & $30.8$ {\small$\pm${0.8}}          & $53.3$ {\small$\pm${0.0}}             & $23.0$  {\small$\pm${0.3}}                \\
            $\times$ &--& IRM                        & $64.2$ {\small$\pm${8.1}}          & $66.3$ {\small$\pm${2.1}}          & $30.0$ {\small$\pm${1.4}}          & $52.4$ {\small$\pm${0.8}}             & $8.2$   {\small$\pm${1.1}}                 \\ 
    \midrule
            ERM &--& ERM                             & $74.3$ {\small$\pm${6.0}}          & $55.5$ {\small$\pm${1.8}}          & $33.6$ {\small$\pm${1.0}}          & $51.1$ {\small$\pm${0.6}}             & $30.2$  {\small$\pm${0.1}}      \\
            ERM &--& GDRO                            & $76.1$ {\small$\pm${6.5}}          & $69.5$ {\small$\pm${0.2}}          & $33.0$ {\small$\pm${0.5}}          & $52.0$ {\small$\pm${0.0}}             & $30.0$  {\small$\pm${0.1}}      \\
            ERM &--& IRM                             & $75.7$ {\small$\pm${7.4}}          & $68.8$ {\small$\pm${1.0}}          & $33.5$ {\small$\pm${1.1}}          & $52.0$ {\small$\pm${0.0}}             & $30.1$  {\small$\pm${0.1}}      \\
    \midrule
            Bonsai &--& ERM                          & $74.0$ {\small$\pm${5.3}}          & $63.3$ {\small$\pm${3.5}}          & $31.9$ {\small$\pm${0.5}}          & $48.6$ {\small$\pm${0.6}}             & $24.2$  {\small$\pm${0.4}}                \\
            Bonsai &--& GDRO                         & $72.8$ {\small$\pm${5.4}}          & $70.2$ {\small$\pm${1.3}}          & $33.1$ {\small$\pm${1.2}}          & $42.7$ {\small$\pm${1.1}}             & $23.0$  {\small$\pm${0.5}}                \\
            Bonsai &--& IRM                          & $73.6$ {\small$\pm${6.2}}          & $68.4$ {\small$\pm${2.0}}          & $32.5$ {\small$\pm${1.2}}          & $47.1$ {\small$\pm${0.6}}             & $23.4$  {\small$\pm${0.4}}                \\
    \midrule
            FeAT &--& ERM                            & $77.8$ {\small$\pm${2.5}}          & $68.1$ {\small$\pm${2.3}}          & $33.1$ {\small$\pm${0.8}}          & $52.9$ {\small$\pm${0.6}}             & $\underline{30.7}$  {\small$\pm${0.4}}    \\
            FeAT &--& GDRO                           & $\underline{80.4}$ {\small$\pm${3.3}}          & $\underline{71.3}$ {\small$\pm${0.5}}          & $33.6$ {\small$\pm${1.7}}          & $52.6$ {\small$\pm${0.6}}             & $30.0$  {\small$\pm${0.1}}    \\
            FeAT &--& IRM                            & $78.0$ {\small$\pm${3.1}}          & $70.3$ {\small$\pm${1.1}}          & $\underline{34.0}$ {\small$\pm${0.7}}          & $52.9$ {\small$\pm${0.6}}             & $30.0$  {\small$\pm${0.2}}    \\
    \midrule
            $\times$ &--& ASGDRO                     & $\mathbf{81.0}$ {\small$\pm${3.8}} & $\mathbf{71.8}$ {\small$\pm${0.4}}   & $\mathbf{35.0}$ {\small$\pm${0.3}} & $\mathbf{54.5}$ {\small$\pm${0.5}}    & $\mathbf{32.2}$ {\small{$\pm${0.2}}} \\
    \bottomrule
        \end{tabular}
\end{adjustbox}
\caption{\textbf{Wilds Benchmark.} Out-of-distribution generalization performances on wilds benchmark with rich representation. The performances of the baseline models are the reported results from \cite{koh2021wilds} and \cite{chen2024understanding}. $\times$ indicates the absence of a pre-training process on the target dataset. Refer to Appendix \ref{append:wilds} for error bars.}
\label{tab:wilds_rich}
\end{table}

\begin{table}[t!] 
\centering
\begin{adjustbox}{width=\linewidth}
\begin{tabular}{lccccc|c}
\toprule
\textbf{Method}  & \textbf{PACS} & \textbf{VLCS} & \textbf{OH} & \textbf{TI} & \textbf{DN} & \textbf{Avg} \\
\midrule
 ERM$^\dag$              & 85.5          & 77.5          & 66.5                & 46.1                    & 40.9               & 63.3         \\
 IRM$^\dag$              & 83.5          & 78.6          & 64.3                & 47.6                    & 33.9               & 61.6         \\
 GDRO$^\dag$             & 84.4          & 76.7          & 66.0                & 43.2                    & 33.3               & 60.7         \\
 I-Mixup$^\dag$          & 84.6          & 77.4          & 68.1                & 47.9                    & 39.2               & 63.4         \\
 MMD$^\dag$              & 84.7          & 77.5          & 66.4                & 42.2                    & 23.4               & 58.8         \\
 SagNet$^\dag$           & 86.3          & 77.8          & 68.1                & \underline{48.6}                    & 40.3               & 64.2         \\
 ARM$^\dag$              & 85.1          & 77.6          & 64.8                & 45.5                    & 35.5               & 61.7         \\
 VREx$^\dag$             & 84.9          & 78.3          & 66.4                & 46.4                    & 33.6               & 61.9         \\
 RSC$^\dag$              & 85.2          & 77.1          & 65.5                & 46.6                    & 38.9               & 62.7         \\
    GSAM \cite{zhuang2021surrogate} & 85.9 & \underline{79.1} & \textbf{69.3} & 47.0 & \underline{44.6} & \underline{65.1} \\  
    RDM \cite{nguyen2024domain} & \textbf{87.2} & 78.4 & 67.3 & 47.5 & 43.4 & 64.8 \\
    RS-SCM \cite{chen2024diagnosing}& 85.8 & 77.6 & 68.8 & 47.6 & 42.5 & 64.4 \\
    LFME \cite{chen2024lfme} & 85.0 & 78.4 & 69.1 & 48.3 & 42.1 & 64.6 \\ 
    ASGDRO  & \underline{86.7} & \textbf{80.0} & \underline{69.2} &\textbf{48.8} & \textbf{44.9} & \textbf{65.9} \\  \midrule
DPLCLIP          & \underline{96.6}          & 79.0          & 82.7                & 45.4                    & \underline{59.1}               & 72.6         \\
DPLCLIP+GDRO     & 95.9          & \underline{79.7}          & \underline{83.6}                & \underline{46.0}                    & \underline{59.1}               & \underline{72.9}         \\
DPLCLIP+ASGDRO   & \textbf{96.8}          & \textbf{80.7}          & \textbf{83.7}               & \textbf{48.9}                    & \textbf{59.8}              & \textbf{74.0}         \\ \bottomrule
\end{tabular}
\end{adjustbox}
\vspace{-0.5em}
\captionof{table}{\small \textbf{DomainBed.} The symbol $^\dag$ indicates reported performance in \citet{gulrajani2020search}. Refer to Appendix \ref{append:domainbed} error bars and experimental details.} \label{tab:dplclip}
\vspace{-1.5em}
\end{table}

We conduct experiments for subpopulation shift, CMNIST \citep{arjovsky2019invariant}, Waterbirds \citep{sagawa2019distributionally}, CelebA \citep{liu2015deep}, and CivilComments \citep{borkan2019nuanced}. The goal of the subpopulation shift task is to obtain the better worst group performance by learning invariant features. Different from H-CMNIST, the spurious correlation acts as a stronger shortcut. As a result, the models cannot learn any invariant feature easily.
\cref{tab:subpop} shows the results of subpopulation shift experiments. 
ASAM, which considers flatness, fails to eliminate spurious correlations and shows limited predictive accuracy on the worst group. On the other hand, ASGDRO shows the best and worst group performance for all data except CivilComments. For CivilComments data, ASGDRO also shows comparable performance with the best algorithms among the baselines. Compared to GDRO, the primary distinction of ASGDRO is its ability to find a common flat minima, which not only enhances robustness for the worst group but also reduces the gap between average accuracy and worst group accuracy. Therefore, \cref{tab:subpop} provides support for our claim that sufficiently learning diverse invariant mechanisms leads to robust generalization performance.

One approach to training a robust model is to enrich the representation learning of invariant features \cite{zhang2022rich,chen2024understanding} rather than training by ERM. This process consists of a pre-training (PT) stage dedicated to representation learning, followed by a fine-tuning (FT) stage utilizing existing invariant learning algorithms. In \cref{tab:wilds_rich}, we compare these algorithms with ASGDRO, evaluated on the Wilds benchmark dataset, which includes various types of distribution shifts collected from real-world scenarios. Notably, the superior performance of ASGDRO, even compared to invariant learning algorithms trained with rich representations during the FT stage, suggests that it is important not only to learn rich representations of invariant features but also to ensure that predictions are composed using diverse invariant features by learning sufficiently diverse invariant mechanisms.

\begin{figure}[t]
\includegraphics[width=\linewidth]{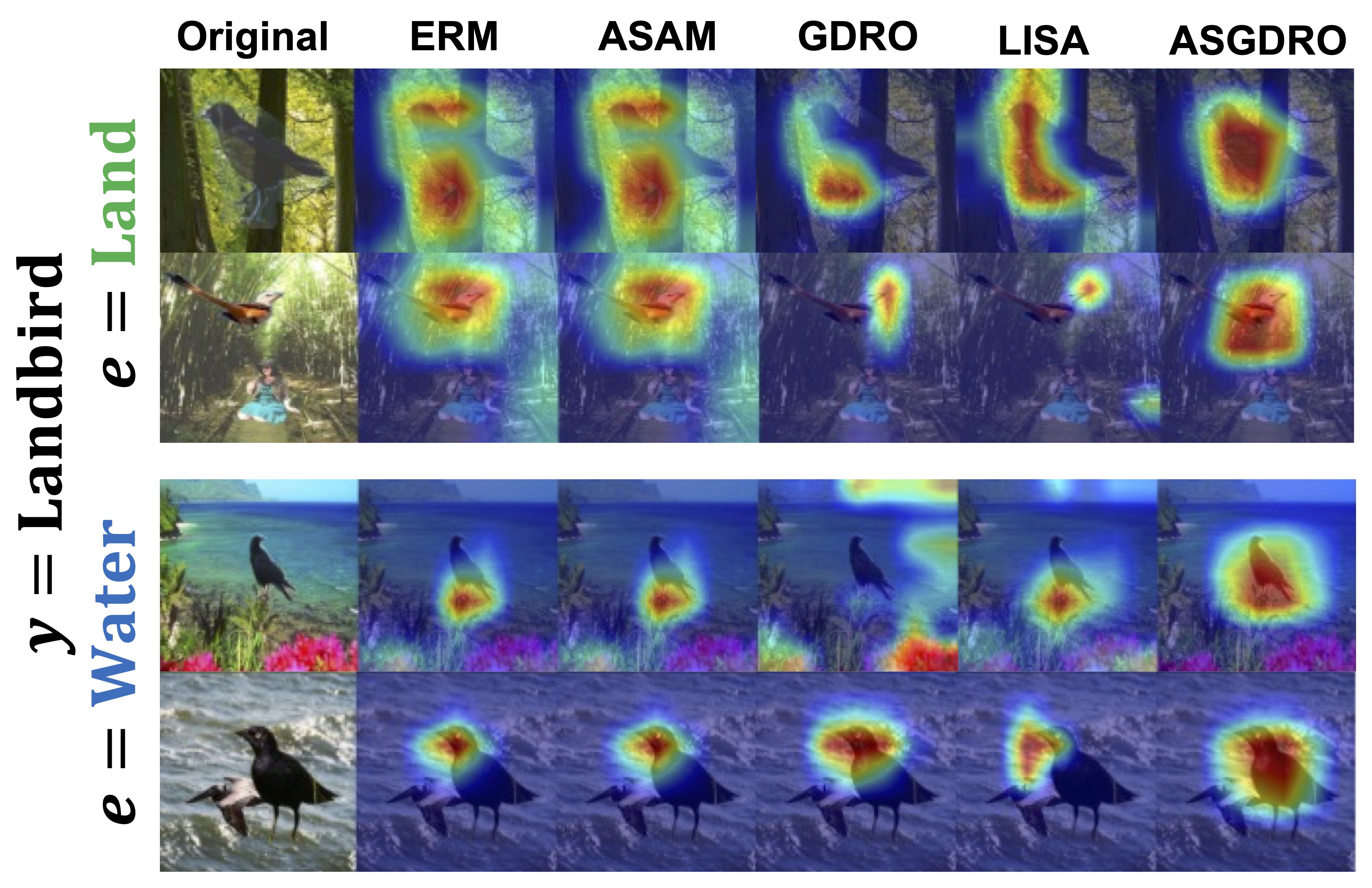}
\vspace{-2em}
\captionof{figure}{\small\textbf{Grad-CAM} ASGDRO learns diverse invariant features.}
\vspace{-1.5em}
\label{fig:camm}
\end{figure}

 We also conduct DomainBed benchmark \citep{gulrajani2020search}, which is the most commonly used for evaluating domain generalization performance under a fair setting. ASGDRO is a model-agnostic method and is easily applied to various algorithms. Thus, we apply ASGDRO with DPLCLIP \citep{zhang2021amortized}, which performs the prompt learning for domain generalization. \cref{tab:dplclip} presents the performance of both the original ASGDRO and DPLCLIP when ASGDRO is applied. ASGDRO achieves the highest average performance compared to other algorithms. Additionally, for DPLCLIP, training with ASGDRO proves to be more effective across all datasets compared to training with standard ERM or GDRO.
 
\subsection{Visual Interpretation by Grad-CAM}\label{sec:gradcam}
    We conduct Grad-CAM analysis to verify whether the effect of learning SIL is being properly applied on the ground-truth label (\cref{fig:camm}). The minority group, land birds on a water background, is underrepresented by the spurious correlation as it has only a few samples. ERM and ASAM use several features to predict the majority group, land birds on a land background, but fail to remove spurious correlation. As a result, they also use the background feature. For the minority groups, however, only a small part of the invariant features is observed to be used for prediction. GDRO successfully removes spurious correlation regardless of the group but still uses only the part of invariant features for prediction. On the other hand, ASGDRO focuses on various invariant features for prediction regardless of the group; that is, it sufficiently uses diverse invariant features of land birds. Additionally, ASGDRO successfully excludes spurious features in their prediction. Appendix \ref{append:gradcam} provides additional results on Grad-CAM.

\subsection{Hessian Analysis}\label{sec:hessian}
\begin{figure}[t]
\centering
\includegraphics[width=0.9\linewidth]{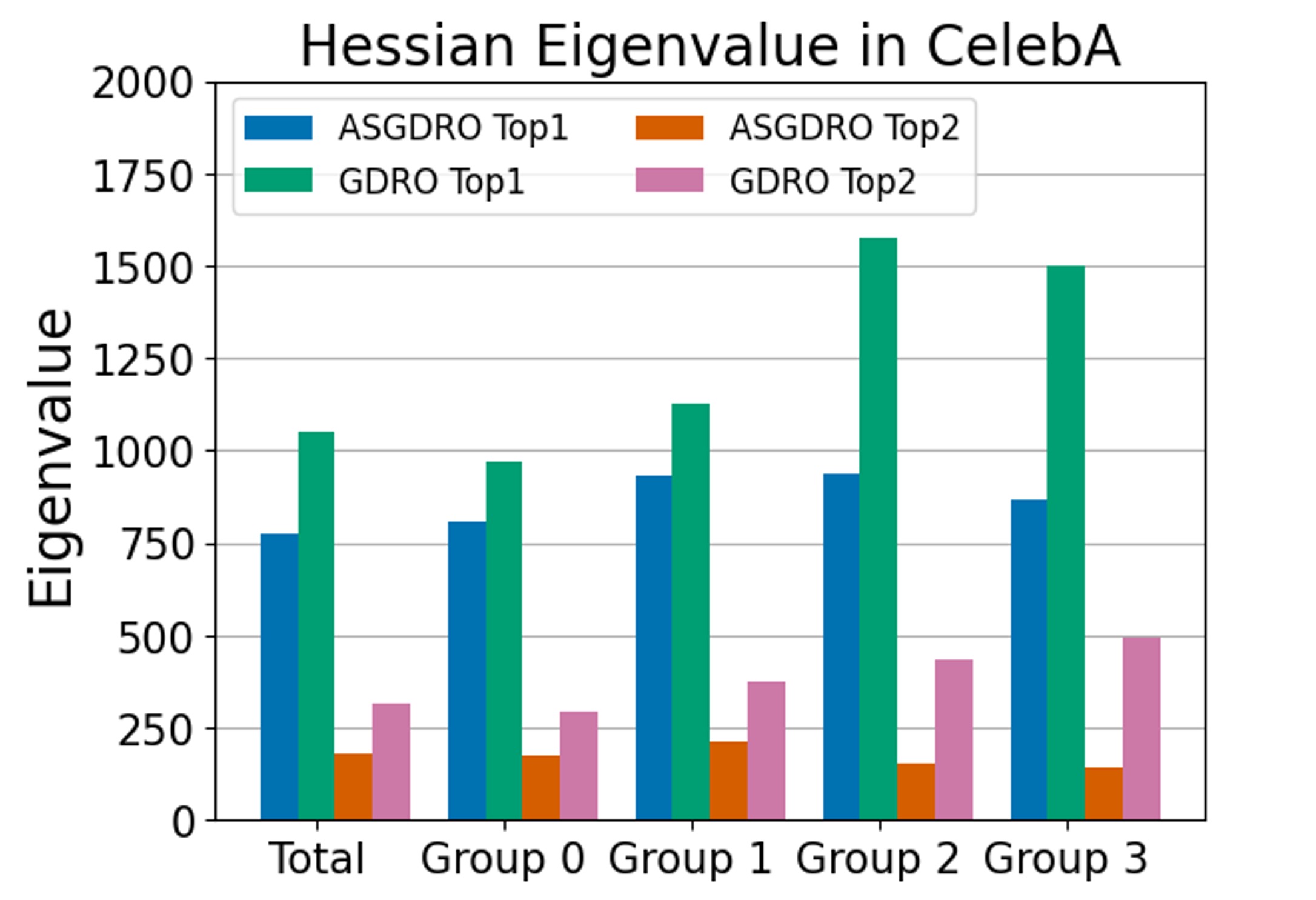}
\vspace{-1em}
\captionof{figure}{\small \textbf{Hessian Analysis on CelebA.} ASGDRO finds the common flat minima for all groups.} \label{fig:hess_celeba}
\vspace{-2em}
\end{figure}

In \cref{fig:hess_celeba}, we report the eigenvalues of the Hessian matrix to measure and compare the flatness of the model \cite{yao2020pyhessian}.  A lower eigenvalue indicates a flatter minima. Compared to GDRO, ASGDRO exhibits lower eigenvalues across all groups. Furthermore, GDRO shows particularly sharper minima in Group 2 and 3, which include minority groups. In contrast, ASGDRO maintains relatively uniform eigenvalues regardless of the group. This suggests that ASGDRO indeed finds a common flat minima, with the regularization for such minima enabling the model to make robust predictions by leveraging diverse invariant mechanisms. Refer to Appendix \ref{append:hess} for additional experimental analysis.

\section{Conclusion} \label{sec5}
This study highlights the significance of SIL, which promotes the learning of diverse invariant features. Unlike invariant learning, SIL enables models to leverage these diverse invariant mechanisms for prediction, ensuring robustness even in environments where some invariant features are unobserved. We also introduce ASGDRO, the first SIL algorithm designed to identify common flat minima across environments. Through both theoretical analysis and experimental validation, we demonstrate that ASGDRO effectively learns diverse invariant mechanisms and finds a common flat minima, which in turn facilitates SIL. We further validate the effectiveness of SIL by demonstrating the generalization capabilities of ASGDRO on our newly developed synthetic SIL dataset, H-CMNIST, as well as on various types of distribution shift benchmark datasets.

\section{Acknowledgement} \label{sec6}
This work was supported by the National Research Foundation of Korea (NRF) grant funded by the Korea government (MSIT) (RS-2024-00457216)
{
    \small
    \bibliographystyle{ieeenat_fullname}
    \bibliography{main}
}
\appendix
\newpage
\onecolumn
\section{Appendix}
\subsection{Limitations and Future Works} \label{append:limit}
ASGDRO utilizes adversarial perturbations to find flat minima, similar to SAM. It requires two forward and backward passes in a single training iteration, which is one of the persistent issues with SAM-based algorithms. However, recent research has been actively focusing on improving the computational cost of SAM \citep{du2021efficient,du2022sharpness}. The computational cost of ASGDRO can also be improved in a similar context, and we consider this to be a future work.

To evaluate whether the algorithm effectively learns diverse invariant mechanisms sufficiently and performs robust predictions, a new benchmark dataset is necessary. Unlike existing invariant learning benchmarks that only require a small number of attributes, constructing an SIL benchmark demands rich attribute annotations to form multiple invariant features. In this paper, we attempt to validate SIL using H-CMNIST, but it is a synthetic dataset based on MNIST. This implies the need for a new benchmark to validate SIL on real-world data, which we leave it as a future work.

\subsection{The subset relationship of invariant features} \label{append:wlog}
In Definition 3, $h_{\theta_{g}}(\hat{Z}^{\text{I}})$ refers to a classifier that relies solely on $\hat{Z}^{\text{I}}\subseteq Z^{\text{I}}$. Given a single sample, if any invariant feature within $\hat{Z}^{\text{I}}$ is observed, we expect the loss evaluated by the classifier to be very small. For two different subset $\hat{Z}^{\text{I}}_{a},\hat{Z}^{\text{I}}_{b}\subseteq\hat{Z}^{\text{I}}$ that satisfy $\hat{Z}^{\text{I}}_b \subseteq \hat{Z}^{\text{I}}_a$, the following inequality holds:
\begin{equation}
    P(\hat{Z}^{\text{I}}_{i} \subseteq\hat{Z}^{\text{I}}_b \text{ is observed in }e\in\mathcal{E}) \leq P(\hat{Z}^{\text{I}}_{i} \subseteq\hat{Z}^{\text{I}}_a \text{ is observed in }e\in\mathcal{E}).  \nonumber
\end{equation}
where $P$ denotes the probability. Note that $Z^{\text{I}}$ also can be partitioned as follows:
\begin{equation}
    Z^{\text{I}} = \bigcup_{i=1}^{p} \{\hat{Z}^{\text{I}}\mid \lvert \hat{Z}^{\text{I}} \rvert=i\},  \nonumber
\end{equation}
where $\lvert \cdot \rvert$ denotes the cardinality of a set and $p$ the number of invariant features. It follows that
\begin{align}
    \max_{\hat{Z}^{\text{I}} \subseteq Z^{\text{I}}} \mathbb{E}[\ell(h_{\theta_h}(\hat{Z}^{\text{I}}),Y^{e})] 
    &=\max \Big[\mathbb{E}[\ell(h_{\theta_h}(Z^{\text{I}}),Y^{e})],  \nonumber  \\
    &\max_{\substack{\hat{Z}^{\text{I}} \subseteq Z^{\text{I}} \\ s.t. \lvert \hat{Z}^{\text{I}} \rvert=p-1}}\mathbb{E}[\ell(h_{\theta_h}(\hat{Z}^{\text{I}}),Y^{e})],  \nonumber  \\
    &\dots,   \nonumber \\ 
    &\max_{\substack{\hat{Z}^{\text{I}} \subseteq Z^{\text{I}} \\ s.t. \lvert \hat{Z}^{\text{I}} \rvert=1}}\mathbb{E}[\ell(h_{\theta_h}(\hat{Z}^{\text{I}}),Y^{e})]\Big]  \nonumber  \\ 
    &=\max_{\substack{\hat{Z}^{\text{I}} \subseteq Z^{\text{I}} \\ s.t. \lvert \hat{Z}^{\text{I}} \rvert=1}}\mathbb{E}[\ell(h_{\theta_h}(\hat{Z}^{\text{I}}),Y^{e})]    \nonumber \\
    &=\max_{Z^{\text{I}}_i \subset Z^{\text{I}}}\mathbb{E}[\ell(h_{\theta_h}(Z^{\text{I}}_i),Y^{e})],   \nonumber
\end{align}
assuming that observing additional invariant features do not adversely affect the performance of the current model.

\subsection{Proof of Proposition 1} \label{append:common_flat}
\begin{proof} \label{append:proof_prop}
Recall that objective function of ASGDRO (Equation 9) is as follows:
\begin{equation}
    \max_{e \in \mathcal{E}}\max_{||\epsilon_{e}|| \leq \rho} \mathcal{R}^{e}(\theta + \epsilon_{e}).   \nonumber
\end{equation}
We use $\mathcal{E}$ instead of $\mathcal{E}_{\text{tr}}$, since this property of ASGDRO holds in any set of environments.
As $\mathcal{R}^{e}(\theta)$ is independent of $\epsilon_{e}$, it can be factored out of the maximization term over $\epsilon_{e}$ as follows:
\begin{equation} \label{eq:decompose_sharpness}
\max_{e \in \mathcal{E}} \max_{||\epsilon_{e}|| \leq \rho} \mathcal{R}^{e}(\theta + \epsilon_{e}) = \max_{e\in \mathcal{E}}[\mathcal{R}^{e}(\theta)+ \max_{||\epsilon_{e}||\leq \rho}[\mathcal{R}^{e}(\theta+\epsilon_{e})-\mathcal{R}^{e}(\theta)]]     \nonumber
\end{equation}
Note that we intentionally add and subtract $\mathcal{R}_{e}$ to reformulate the expression, enabling the separation of terms for clearer analysis.
Using the Taylor approximation expanded up to the first-order term, we have:
\begin{align}
    \max_{e\in \mathcal{E}}[\mathcal{R}^{e}(\theta)+ \max_{||\epsilon_{e}||\leq \rho}[\mathcal{R}^{e}(\theta+\epsilon_{e})-\mathcal{R}^{e}(\theta)]] &\approx \max_{e\in \mathcal{E}}[\mathcal{R}^{e}(\theta)+ \max_{||\epsilon_{e}||\leq \rho}[\epsilon_{e} \cdot \nabla \mathcal{R}^{e}(\theta)]]  \nonumber  \\
    & = \max_{e\in \mathcal{E}}[\mathcal{R}^{e}(\theta)+ \epsilon_{e}^{*} \cdot \nabla \mathcal{R}^{e}(\theta)], \label{eq:max}
\end{align}
where $\epsilon_{e}^{*}= \rho\frac{\nabla\mathcal{R}^{e}(\theta)}{||\nabla\mathcal{R}^{e}(\theta)||}$. Note that Eq. \ref{eq:max} holds because the maximum value over $||\epsilon_{e}||\leq\rho$ is achieved when $\epsilon_{e}$ and $\nabla\mathcal{R}^{e}(\theta)$ are aligned in the same direction \citep{foret2020sharpness}. By substituting $\epsilon^{*}_{e}$, we obtain the following equation:
\begin{equation}
\max_{e\in \mathcal{E}}[\mathcal{R}^{e}(\theta)+ \epsilon_{e}^{*} \cdot \nabla \mathcal{R}^{e}(\theta)] = \max_{e\in \mathcal{E}}[\mathcal{R}^{e}(\theta)+ \rho|| \nabla\mathcal{R}^{e}(\theta) ||].
\label{eq:gradientnorm}
\end{equation}
\citet{zhao2022penalizing} demonstrate that minimizing the gradient norm of the risk leads to finding flat minima. Eq. \ref{eq:gradientnorm} minimizes both risk and the gradient norm of risk for each environment. Consequently, ASGDRO constrains the training process to find a common flat minimum across environments.

\end{proof}

\subsection{Proof of Theorem 1} \label{append:sil}
\begin{proof} \label{append:proof_theorem}
In this setting, we consider a single input for each environment $e$. Suppose there are $p$ invariant features, and every invariant feature has the same activation:
\begin{equation}
    Z^{\text{I}} = (1, \dots, 1),  \nonumber  
\end{equation}
where $\lvert Z^{\text{I}} \rvert = p$. We assume that all spurious features are completely removed. Thus, $Z = (Z^{\text{I}}, Z^{\text{NI}}) = Z^{\text{I}}$, where $\lvert Z \rvert = p$. Consequently, the risk for $Z$ is identical across all environments $e$:
\begin{equation} \label{eq:equal}
    \mathcal{R}^{e}(\theta) = \mathcal{R}^{e'}(\theta)=c \quad \text{for any $e,e'\in\mathcal{E}_{\text{tr}}$,}
\end{equation}
where $c$ is a constant. Given $Z^{\text{I}}$, we focus only on the parameters of the classifier, denoted by $\theta^{\text{I}}$. Recall that the classifier satisfying Eq. 3 in main paper, and Eq. \ref{eq:equal}, is not unique. Define $\theta^{\text{I}}_i$ as the classifier that utilizes only the $i$-th element of $Z^{\text{I}}$.

For simplicity, let $\theta^{\text{I}}_{i}$ be a column vector where only the $i$--th element is one, and all other elements are zero:

\begin{equation}
    Z^{\text{I}}\theta^{\text{I}}_{i}=Z^{\text{I}}_{i}=1.    
\end{equation}
Furthermore, the convex combination of $\theta^{\text{I}}_i$ also yields an equivalent output:
\begin{equation}
    Z^{\text{I}}\sum^{p}_{i=1}\lambda_i\theta^{\text{I}}_{i}=1,    \nonumber
\end{equation}
where $\sum_{i=1}^p \lambda_i = 1$ and $0 \leq \lambda_i \leq 1$ for all $i \in \{1, \dots, p\}$. We denote the current classifier as $\theta^{\text{I}}_\lambda := \sum_{i=1}^p \lambda_i \theta^{\text{I}}_i$, where $\lambda = (\lambda_1, \dots, \lambda_p)$. From Proposition 1, we know:
\begin{equation} \label{eq:goal}
    \max_{e \in \mathcal{E}}\max_{||\epsilon_{e}|| \leq \rho} \mathcal{R}^{e}(\theta + \epsilon_{e})=\max_{e \in \mathcal{E}_{\text{tr}}} \left[ \mathcal{R}^{e}(\theta)+\rho||\nabla_{\theta}\mathcal{R}^{e}(\theta)|| \right].
\end{equation}
For the mean-squared error loss function $\mathcal{R}^e(\theta) = \frac{1}{2} \|Y^e - \sum_{i=1}^p \theta_i\|^2$, the gradient is given by $\nabla \mathcal{R}^e(\theta) = -(Y^e - \sum_{i=1}^p \theta_i) \cdot \boldsymbol{1}$, where $\boldsymbol{1}$ is a $p$-dimensional vector whose elements are all equal to 1. Substituting $\theta^{\text{I}}_\lambda$ into Eq. \ref{eq:goal}, we get:
\begin{equation}
    \max_{e \in \mathcal{E}} \max_{\|\epsilon_e\| \leq \rho} \mathcal{R}^e(\theta^{\text{I}}_\lambda + \epsilon_e) = \max_{e \in \mathcal{E}_{\text{tr}}} \left[ \mathcal{R}^e(\theta^{\text{I}}_\lambda) + \rho \|\nabla_\theta \mathcal{R}^e(\theta^{\text{I}}_\lambda)\| \right].    \nonumber
\end{equation}
This simplifies to:
\begin{equation}
    \max_{e \in \mathcal{E}_{\text{tr}}} \left[ \mathcal{R}^e(\theta^{\text{I}}_\lambda) + \rho \|-\lambda \odot \nabla \mathcal{R}^e(\theta^{\text{I}}_\lambda)\| \right] = \max_{e \in \mathcal{E}_{\text{tr}}} \left[ \mathcal{R}^e(\theta^{\text{I}}_\lambda) + \rho \|\lambda\| \cdot \|\nabla \mathcal{R}^e(\theta^{\text{I}}_\lambda)\| \right],    \nonumber
\end{equation}
where $\mathcal{R}^e(\theta^{\text{I}}_\lambda) = c$ for any $\lambda$. Since the classifier uses only invariant features, minimizing the adversarial term reduces to:
\begin{align}
    \argmin_{\lambda}\max_{e \in \mathcal{E}_{\text{tr}}} \max_{||\epsilon|| \leq \rho} \mathcal{R}^{e}(\theta^{\text{I}}_\lambda + \epsilon) &= \argmin_{\lambda}\max_{e \in \mathcal{E}_{\text{tr}}} \left[ \mathcal{R}^{e}(\theta^{I}_{\lambda}) +\rho||\lambda|| \cdot ||\nabla\mathcal{R}^{e}(\theta^{I}_{\lambda})|| \right]   \nonumber  \\ 
    &=\argmin_{\lambda} ||\lambda||.    \nonumber
\end{align}
By the Cauchy-Schwarz inequality:
\begin{equation}
    \left( \sum_{i=1}^{p} \lambda_i \right)^{2} \leq p \cdot \sum_{i=1}^{p} \lambda_i^{2} = p\cdot||\lambda||^{2}.    \nonumber
\end{equation}
Under the condition $\sum^{p}_{i=1}\lambda_i=1$, equality holds when $\lambda_i = \frac{1}{p}$ for all $i$, yielding:
\begin{equation}
    \argmin_{\lambda} ||\lambda|| = (\frac{1}{p}, \dots , \frac{1}{p})    \nonumber
\end{equation}
\end{proof}

\subsection{Mechanism of ASGDRO for Removing Spurious Features} \label{append:spurious}
ASGDRO successfully removes spurious features. Inspired by \citet{andriushchenko2023sharpness}, we reformulate the two-layer ReLU case presented in that paper to demonstrate this. Consider a two--layer ReLU network
\begin{equation}
    f(\theta) = \langle \theta_h, \sigma(\theta_g x) \rangle, \nonumber
\end{equation}
where $\theta=(\theta_g,\theta_h)$, $\theta_g\in\mathbb{R}^{k\times m}$ and $\theta_h\in\mathbb{R}^{k}$. Recall that ASGDRO minimizes the maximum sharpness across environments:
\begin{equation}
    \max_{e \in \mathcal{E}}\max_{||\epsilon_{e}|| \leq \rho} \mathcal{R}^{e}(\theta + \epsilon_{e}). \nonumber
\end{equation}
Let $e_t$ denote the environment that attains the maximum risk at the current step $t$. Then, the adversarial perturbation is $\epsilon_{e_{t}}^{*}=\rho\frac{\nabla\mathcal{R}^{e_{t}}(\theta)}{\|\nabla\mathcal{R}^{e_{t}}(\theta)\|}$ \citep{foret2020sharpness} and the risk is
\begin{equation}
   \max_{||\epsilon_{e_{t}}|| \leq \rho} \mathcal{R}^{e_{t}}(\theta + \epsilon_{e_{t}})= \mathcal{R}^{e_{t}}(\theta + \rho\frac{\nabla\mathcal{R}^{e_{t}}(\theta)}{\|\nabla\mathcal{R}^{e_{t}}(\theta)\|})\nonumber
\end{equation}
Under the first--order Taylor approximation,
\begin{equation}
    \nabla \mathcal{R}^{e_{t}} \left( \theta + \rho\frac{\nabla\mathcal{R}^{e_{t}}(\theta)}{\|\nabla\mathcal{R}^{e_{t}}(\theta)\|}\right) \approx \nabla \left[ \mathcal{R}^{e_{t}}(\theta) + \rho \| \nabla\mathcal{R}^{e_{t}}(\theta)\|\right] \nonumber
\end{equation}
\cite{andriushchenko2023sharpness} shows that under two--layer ReLU network, the update rule for pre-activation of k--th neuron is as follows:
\begin{align}
    \langle \theta_{g}^{(k)}, x \rangle ^{(t+1)} \approx \langle \theta_{g}^{(k)}, x \rangle ^{(t)} &\underbrace{- \eta \gamma \left( 1+\rho \frac{\|\nabla f(\theta)\|}{\sqrt{\mathcal{R}^{e_{t}}(\theta)}} \right)a_{k}\sigma'(\langle \theta_{g}^{(k)}, x  \rangle)\|x\|^{2}}_{\text{(a)}} \nonumber\\
    &\underbrace{- \eta \rho \frac{\sqrt{\mathcal{R}^{e_{t}}(\theta)}}{\|\nabla f(\theta)\|} \sigma(\langle \theta_{g}^{(k)}, x  \rangle)\|x\|^{2}}_{\text{(b)}}, \nonumber
\end{align}
where $\eta$ denotes the learning rate, $\gamma=f(\theta)-y$, i.e. the residual. 

In ASGDRO, regularization on the gradient norm has two key effects. 
First, as seen in term (a), the gradient update direction remains the same, but the model is updated with a larger learning rate. 
Second, in term (b), when $\mathcal{R}^{e_{t}}(\theta)$ is large enough, the pre-activation of the $k$-th neuron, $\langle \theta_{g}^{(k)}, x \rangle$, turns negative. Note that a large $\mathcal{R}^{e_{t}}$ implies that highly activated neurons at this point tend to encode significant information from spurious features. When $\mathcal{R}^{e_{t}}(\theta)$ causes the pre-activation of a neuron to become negative, the nature of the ReLU activation function ensures that the output of that neuron becomes zero. 
This indicates that, under distribution shifts, regularization via the common flat minima in ASGDRO effectively removes spurious features.

\subsection{Heterogeneous-CMNIST (H-CMNIST)} \label{append:hcmnist}
\subsubsection*{Dataset Details}
The test set of H-CMNIST is constructed by flipping the proportion of $Z_{\text{BP}}$ from the training set. H-CMNIST conducts two types of tests. First, TestBed 1 evaluates whether the algorithm learns at least one invariant feature. To assess this, it compares the prediction differences between cases where the spurious feature $Z_{\text{BP}}$ is present and absent. TestBed 2 examines whether the model remains robust to $Z_{\text{BP}}$ and maintains good performance when only $Z_{\text{shape}}$ is present, excluding $Z_{\text{color}}$ among the two invariant features.

\subsubsection*{Experimental Details}
In H-CMNIST experiments, we use ResNet18 \citep{he2016deep} with SGD. We also conduct reweighted sampling when the algorithm setting can use the environment information, i.e., GDRO \citep{sagawa2019distributionally} and ASGDRO. In the H-CMNIST experiment, we set the loss of GDRO and ASGDRO by the group, not the domain. That is, there is four groups; (Class=0,BP=Top Left), (Class=0,BP=Bottom Right), (Class=1,BP=Top Left), (Class=1,BP=Bottom Right). For hyperparameter tuning, we perform grid search over learning rate, $\{10^{-3}, 10^{-4}\}$, and $L_2$--regularization, $\{1, 10^{-1}, 10^{-3},10^{-4}\}$. We fix the batch size, 128, and train the model up to 20 epochs. For ASAM \citep{kwon2021asam} and ASGDRO, we search the hyperparameter $\rho$ among $\{0.05, 0.2, 0.5, 0.8\}$. We fix the robust step size, $\gamma$, as 0.01 for GDRO and ASGDRO. We evaluate the models with three random seeds. 

\subsection{Subpopulation Shifts: Datasets and Experimental Details} \label{append:subpop}

\subsubsection*{Dataset Details}
\begin{figure*}[t]
\centering
        \includegraphics[width=0.9\linewidth]{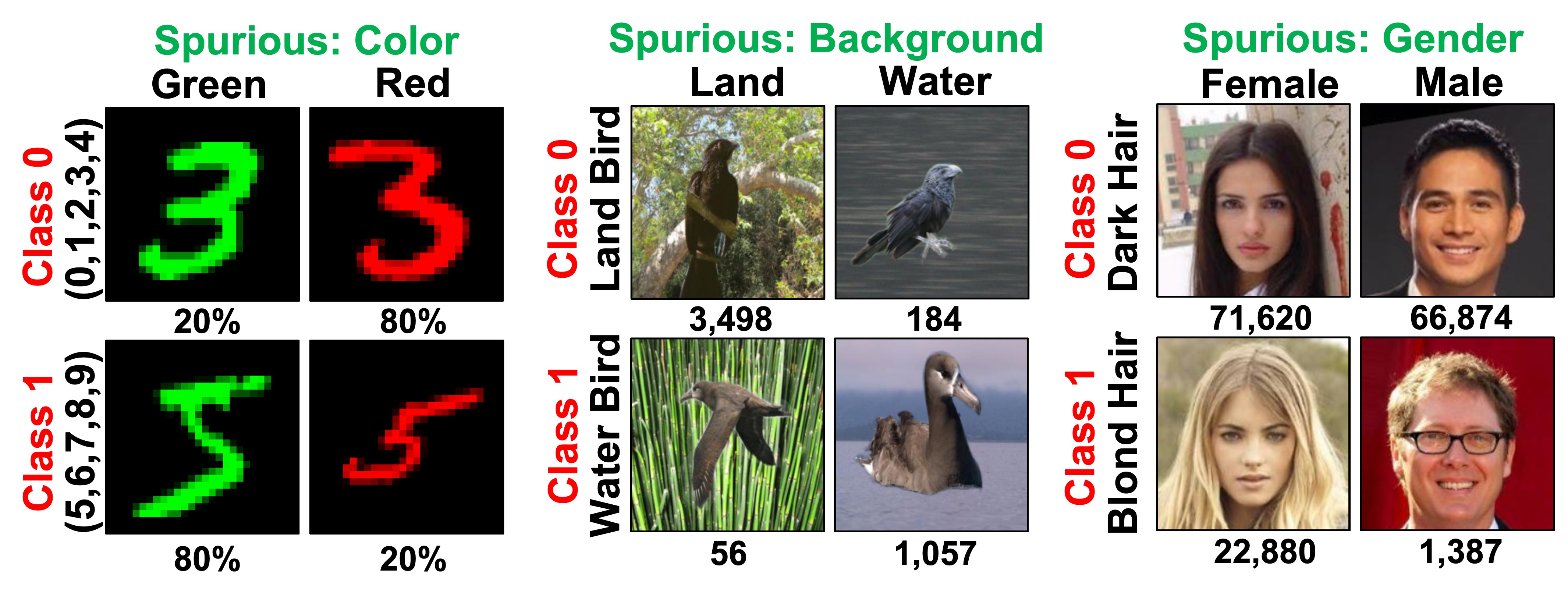}
    \caption{\textbf{CMNIST, Waterbirds, CelebA.} In each dataset, each row represents the class and each column represents the spurious feature. The numbers written below the images represent the ratio or count of data belonging to each group in the training dataset, where each group consists of (Class, Spurious Feature) pairs.}
    \vspace{-1em}
    \label{fig:subpop_dataset}
\end{figure*}

  In Table 2 in the main paper, we conduct our experiment for subpopulation shifts with five datasets: CMNIST \citep{arjovsky2019invariant}, Waterbirds \citep{sagawa2019distributionally}, CelebA \citep{liu2015deep}, CivilComments \citep{borkan2019nuanced}. CMNIST, Waterbirds, and CelebA datasets correspond to computer vision tasks (Figure \ref{fig:subpop_dataset}), while CivilComments pertain to natural language processing tasks. In this section, we will describe each dataset and provide experimental details. To implement this, we utilized the codes provided by \citep{yao2022improving}\footnote{https://github.com/huaxiuyao/LISA}.

\subsubsection*{Colored MNIST (CMNIST)}
In the CMNIST dataset provided by \citep{arjovsky2019invariant}, we perform binary classification to predict which number corresponds to the shape of a given digit. Specifically, when the shape of the digit corresponds to a logit between 0 and 4, the class is assigned as 0, and when it falls between 5 and 9, the class is assigned as 1. However, unlike the original MNIST dataset \citep{lecun1998gradient}, CMNIST introduces color as a spurious feature in the training set. When this spurious correlation becomes stronger than the invariant relationship between the class and the shape of the digit, a model trained without any regularization may be prone to relying on the spurious feature for predictions.

While \citet{arjovsky2019invariant} constructs two environments with different ratios of spurious features in the training set, \citet{yao2022improving} uses a single environment to compose the training set. Our CMNIST dataset experiment follows the same setting as \citep{yao2022improving}, where the dataset consists of four groups when considering combinations of ``Shape of Logit'' and ``Color'' as a single group. Concretely, Class 0 and Class 1 have similar numbers of data points, but the distribution of spurious features differs between the two classes. Class 0 consists of 80\% red logits and 20\% green logits, while Class 1 has 80\% green logits and 20\% red logits. Furthermore, within each class, 25\% of the data acts as label noise, having a logit shape that does not correspond to its class. Therefore, the spurious feature, color, forms a stronger correlation between classes compared to that of the invariant feature, the shape of logits.

The validation set is constructed with an equal number of instances per group. The worst-group accuracy, defined as the lowest accuracy among all the groups, is utilized to select the best model. For the test set, we assume a distribution of the spurious feature that is opposite to the training set. Specifically, for Class 0, 90\% of the data has a red color, and 10\% has a green color, while for Class 1, it is the opposite. It is done to assess whether the model relies on the spurious feature for predictions.

\subsubsection*{Waterbirds}
Waterbirds dataset, constructed by \citep{sagawa2019distributionally}, is designed for the task of determining whether a bird belongs to the Landbird or Waterbird class. It consists of images of birds, from \citep{wah2011caltech}, as the invariant feature, while the spurious feature is the background, from \citep{zhou2017places}, which can either be Water or Land background. Indeed, in the Waterbirds dataset, the groups are formed by the combination of ``Bird'' and ``Background''. Specifically, the bird images corresponding to each class consist of more than 10 different species of birds. On the other hand, each background is composed of two categories obtained from \citep{zhou2017places}. As can be seen in Figure \ref{fig:subpop_dataset}, the Landbird class predominantly has images with Land background, while the majority of images in the Waterbird class have Water background. Therefore, the spurious feature, background, may indeed form a strong spurious correlation with each class.

We follow the setting of previous research, \citep{sagawa2019distributionally,yao2022improving}, for the validation and test processes as well. The best model is selected based on the highest worst-group accuracy on the validation set. Unlike the training set, the validation and test sets are designed to have an equal number of images for each group within each class. When reporting the average accuracy on the test dataset using the best model, we first compute the group accuracy for each group in the test set. Then, we calculate the weighted average of these accuracies using the group distribution from the training set. This approach is adopted to mitigate the uncertainty in estimating group accuracies, as the number of images belonging to the minority group in the Waterbird dataset is significantly smaller compared to other datasets \citep{sagawa2019distributionally}.

\subsubsection*{CelebA}

CelebA dataset by \citep{liu2015deep} is a collection of facial images of celebrities from around the world. It includes attribute values associated with each individual, such as hair color and gender. In order to evaluate the effects of subpopulation shifts, \citet{sagawa2019distributionally} reformulated the CelebA dataset to align with the task of predicting whether the hair color is blond or not. In this case, the spurious feature is gender, and thus, the dataset is composed of four groups based on the combinations of hair color and gender. It can be observed from Figure \ref{fig:subpop_dataset} that images belonging to Class 0, corresponding to dark hair rather, are plentiful regardless of gender. However, for images in Class 1, which represent blond hair, the majority of them are distributed in the Female group. Therefore, gender can act as a spurious feature, and the goal of this task is to obtain a model that focuses solely on the invariant feature, hair color, rather than the face which may capture the characteristics of gender-related features.

The best model is selected based on the best worst-group accuracy on the validation set. In this case, the validation set and test set have the same distribution of images per group as the training set. Therefore, the average test accuracy reflects this distribution accordingly.

\subsubsection*{CivilComments}

The CivilComments dataset, \citep{borkan2019nuanced}, is a dataset that gathers comments from online platforms and is used for the task of classifying whether a given comment is toxic or not. We conduct the experiment on the CivilComments dataset, which has been reformulated by \citep{koh2021wilds}. Each comment is labeled to indicate whether it mentions the presence of any word of the 8 demographic identities; Black, White, Christian, Muslim, other religions, Male and Female. Therefore, the CivilComments dataset consists of 16 groups, formed by the combination of toxic labels and the presence or absence of the 8 demographic identities in each comment. Each demographic identity can potentially act as a spurious feature. To prevent this, the goal of the task is to train the model to focus solely on the invariant feature of toxic labels and not rely on demographic identities as predictive factors.

However, in reality, unlike other datasets, each comment in the CivilComments dataset can mention more than one demographic identity. Considering all possible combinations of demographic identities for each comment and training the model on all these combinations would be inefficient. Therefore, we follow the learning approach proposed by \citep{koh2021wilds}. Concretely, we only consider four groups based on whether the comment mentions toxicity and whether it mentions the demographic identity of being ``Black'', without considering other demographic identities. We train the model using these four groups. However, during the validation and test, we evaluate the model's performance individually for all 16 groups and record the lowest accuracy among the group accuracies as the worst-group accuracy. The Best model is selected based on this worst-group accuracy.

\subsubsection*{Experimental Details}

\begin{table}[t] 
  \centering
  \adjustbox{max width=\linewidth}{
    \begin{tabular}{ccccccccc}
    \toprule
    \multicolumn{1}{c}{} &  \multicolumn{2}{c}{{CMNIST}} &\multicolumn{2}{c}{{Waterbirds}} & \multicolumn{2}{c}{{CelebA}} & \multicolumn{2}{c}{{CivilComments}}  \\
    \multicolumn{1}{c}{}&\multicolumn{1}{c}{Avg} & \multicolumn{1}{c}{Worst} & \multicolumn{1}{c}{Avg} & \multicolumn{1}{c}{Worst} & \multicolumn{1}{c}{Avg} & \multicolumn{1}{c}{Worst} & \multicolumn{1}{c}{Avg} & \multicolumn{1}{c}{Worst}  \\
    \midrule
     ERM$^\ddag$   &  27.8$\pm$ 1.9\%  & 0.0$\pm$ 0.0\% & {\underline{97.0}$\pm$ \underline{0.2\%}}  & 63.7$\pm$ 1.9\%  &  {\textbf{94.9}$\pm$ \textbf{0.2\%}}  & 47.8$\pm$ 3.7\%  & {\underline{92.2}$\pm$ \underline{0.1\%}}  & 56.0$\pm$ 3.6\%   \\
     ASAM  &  40.5$\pm$ 0.8\%  & 34.1$\pm$ 1.2\%  &  {\textbf{97.4}$\pm$ \textbf{0.0\%}}  & 72.4$\pm$ 0.4\%  & 93.7$\pm$ 0.8\%  & 46.5$\pm$ 10.3\%  &  {\textbf{92.3}$\pm$ \textbf{0.1\%}} & 58.9$\pm$ 1.7\% \\
     IRM$^\ddag$ &   72.1$\pm$ 1.2\% & 70.3$\pm$ 0.8\% & 87.5$\pm$ 0.7\% & 75.6$\pm$ 3.1\% & {\underline{94.0}$\pm$ \underline{0.4\%}} & 77.8$\pm$ 3.9\% & 88.8$\pm$ 0.7\% & 66.3$\pm$ 2.1\% \\
     IB-IRM$^\ddag$ &  72.2$\pm$ 1.3\%  & 70.7$\pm$ 1.2\% & 88.5$\pm$ 0.6\% & 76.5$\pm$ 1.2\% & 93.6$\pm$ 0.3\% & 85.0$\pm$ 1.8\% & 89.1$\pm$ 0.3\% & 65.3$\pm$ 1.5\%  \\
     V-REx$^\ddag$ &  71.7$\pm$ 1.2\%  & 70.2$\pm$ 0.9\% & 88.0$\pm$ 1.0\% & 73.6$\pm$ 0.2\% & 92.2$\pm$ 0.1\% & 86.7$\pm$ 1.0\% & 90.2$\pm$ 0.3\% & 64.9$\pm$ 1.2\%  \\
     CORAL$^\ddag$ &  71.8$\pm$ 1.7\%  & 69.5$\pm$ 0.9\% & 90.3$\pm$ 1.1\% & 79.8$\pm$ 1.8\% & 93.8$\pm$ 0.3\% & 76.9$\pm$ 3.6\% & 88.7$\pm$ 0.5\% & 65.6$\pm$ 1.3\% \\
     GDRO$^\ddag$ &  72.3$\pm$ 1.2\%  & 68.6$\pm$ 0.8\% & 91.8$\pm$ 0.3\% & {\underline{90.6}$\pm$ \underline{1.1\%}} & 92.1$\pm$ 0.4\% & 87.2$\pm$ 1.6\% & 89.9$\pm$ 0.5\% & 70.0$\pm$ 2.0\%  \\
     DomainMix$^\ddag$ &  51.4$\pm$ 1.3\%  & 48.0$\pm$ 1.3\% &  76.4$\pm$ 0.3\% & 53.0$\pm$ 1.3\% & 93.4$\pm$ 0.1\% & 65.6$\pm$ 1.7\% & 90.9 $\pm$ 0.4\% & 63.6$\pm$ 2.5\% \\
     Fish$^\ddag$ &  46.9$\pm$ 1.4\%  & 35.6$\pm$ 1.7\% & 85.6$\pm$ 0.4\% & 64.0$\pm$ 0.3\% & 93.1$\pm$ 0.3\% & 61.2$\pm$ 2.5\% & 89.8$\pm$ 0.4\% & 71.1$\pm$ 0.4\%  \\
     LISA$^\ddag$ &  {\underline{74.0}$\pm$ \underline{0.1\%}}  & {\underline{73.3}$\pm$ \underline{0.2\%}} & 91.8$\pm$ 0.3\% & 89.2$\pm$ 0.6\% & 92.4$\pm$ 0.4\% & {\underline{89.3}$\pm$ \underline{1.1\%}} & 89.2$\pm$ 0.9\% &  {\textbf{72.6}$\pm$ \textbf{0.1\%}}  \\
     PDE$^{\ddag\ddag}$         &  --\%  & --\%                 & 92.4$\pm$ 0.8\% & 90.3$\pm$ 0.3\%       & 92.0$\pm$ 0.6\% & \textbf{91.0$\pm$ 0.4\%}& 86.3$\pm$ 1.7\% & 71.5$\pm$ 0.5\%  \\
    \midrule
     ASGDRO &   {\textbf{74.8}$\pm$ \textbf{0.1\%}}  &  {\textbf{74.2}$\pm$ \textbf{0.0\%}} & 92.3$\pm$ 0.1\% &  {\textbf{91.4}$\pm$ \textbf{0.1\%}} & 92.1$\pm$ 0.4\% &  {\textbf{91.0}$\pm$ \textbf{0.5\%}} & 90.2$\pm$ 0.2\% & {\underline{71.8}$\pm$ \underline{0.4\%}}  \\ 
    \bottomrule
    \end{tabular}}
    \vspace{1em}
\caption{\small \textbf{Subpopulation Shift}. $\ddag$ denotes the performance reported from \citep{yao2022improving}, and $\ddag\ddag$ denotes the performance reported from \citep{deng2024robust}. Avg. denotes average accuracy, and Worst denotes worst group accuracy}
\end{table}

The search range of the hyperparameter $\rho$, which determines the range for exploring the flat region, is fixed to $\{0.05, 0.2, 0.5, 0.8, 1.0, 1.2, 1.5\}$ for all datasets. We evaluate the model across three random seeds and report the average performance. We set robust step size $\gamma$, in Algorithm 1 of the main paper, $\{0.1, 0.01\}$. In addition, we use the same range for adjusted-group coefficient $C$, $\{0,1,2,3,4,5\}$ (Section 3.3 in \citep{sagawa2019distributionally} for details). In CMNIST, Waterbirds, and CelebA datasets, we utilize ResNet50 \citep{he2016deep} models. The same hyperparameter ranges are applied to ASAM and ASGDRO, and the other performances for other baselines are reported performances from \citep{liu2021just,yao2022improving,han2022umix}. All experiments in this paper were conducted using NVIDIA RTX A6000 with 49140 MiB of GPU memory and GeForce RTX 3090 with 24.00 GiB of GPU memory.

In CMNIST, we have the same hyperparameter search range as \citep{yao2022improving} by default: batch size 16, learning rate $10^{-3}$, $L_2$--regularization $10^{-4}$ with SGD over 300 epochs. For Waterbirds, we perform the grid search over the batch size, $\{16, 64\}$, the learning rate, $\{10^{-3}, 10^{-4}, 10^{-5}\}$, and $L_2$--regularization, $\{10^{-4}, 10^{-1}, 1\}$. We train our model with SGD over 300 epochs. We also conduct grid search over the batch size, $\{16, 128\}$, the learning rate, $\{10^{-4}, 10^{-5}\}$, and $L_2$--regularization, $\{10^{-4}, 10^{-2}, 1\}$ for CelebA, training with SGD over 50 epochs. We referenced \citep{yao2022improving,liu2021just} for this range of hyperparameter search. For CivilComments, we use DistilBERT \citep{sanh2019distilbert} model. We follow the hyperparameter search range provided in \citep{koh2021wilds}. For optimizer, we use AdamW \citep{loshchilov2017decoupled} with $10^{-2}$ for $L_2$--regularization. We find the optimal learning rate among $\{10^{-6}, 2 \times 10^{-6}, 10^{-5}, 2 \times 10^{-5}\}$. We train up to 5 epochs with batch size 16. The gradient clipping is applied only during the second step, which is the actual update step, in the SAM-based algorithm \citep{foret2020sharpness}. 

\subsection{Error bars for Wilds Benchmark} \label{append:wilds}
\begin{figure}[h!]
\centering
    \includegraphics[width=\textwidth]{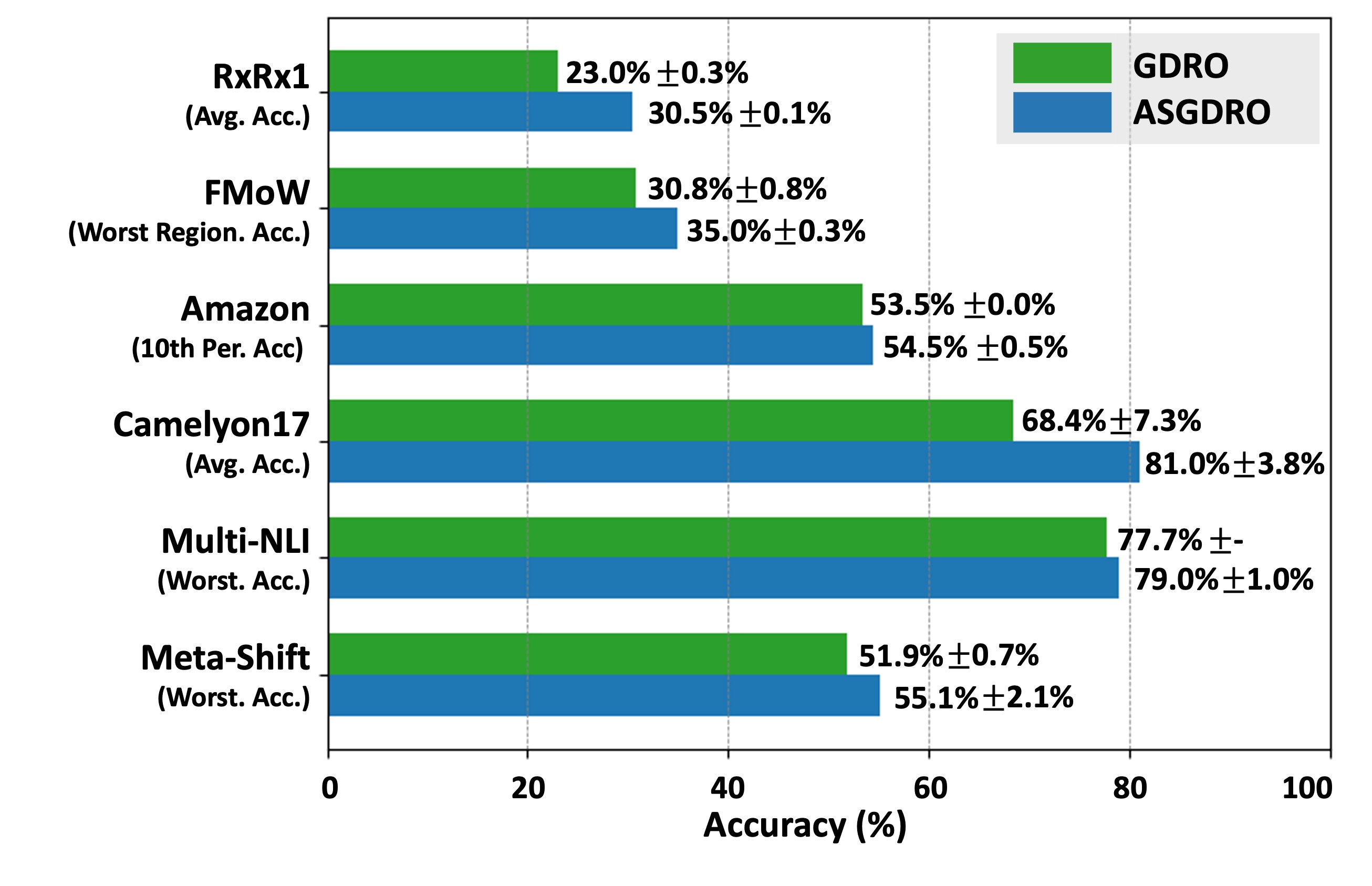}
    \caption{\small \textbf{Standard Deviations for Wilds Benchmark Datasets.}}
    \label{fig:wild_std}
\end{figure}
We demonstrate the differences between GDRO and ASGDRO in various distribution shift scenarios that could occur in the real world. Wilds benchmark \cite{koh2021wilds} consists of datasets collected from the real world. Camelyon17 and RxRx1 are datasets where domain shift is predominant. Amazon and FMoW are datasets where both subpopulation shift and domain shift are simultaneously predominant. Figure \ref{fig:wild_std} shows the results of ASGDRO and GDRO on Wilds Benchmark, MetaShift dataset, and Multi-NLI \citep{williams2017broad}. ASGDRO shows superior performances consistently compared with GDRO. It implies that identifying common flat minima across environments enhances the robustness of models.

\subsection{Experimental Details and Error bars for Domainbed with DPLCLIP} \label{append:domainbed}

\subsubsection*{Experimental Details for DomainBed Experiment}
Using DomainBed framework \citep{gulrajani2020search}, we evaluate domain generalization algorithms by randomly sampling hyperparameter combinations within predefined hyperparameter search ranges for each algorithm. The goal of domain generalization is to train models that perform robustly on unseen domains. Consequently, the choice of the best model is heavily influenced by whether the validation set used for model selection is sampled from the test domain or the train domains. To account for this, we provide results for both the training-domain validation set, which does not utilize information from the test domain, and the test-domain validation set, where model selection is performed using information from the test domain. The following subsections present the results for each dataset, considering both model selection methods.

By combining ASGDRO with the existing successful domain generalization approach, DPLCLIP \citep{zhang2021amortized}\footnote{https://github.com/shogi880/DPLCLIP}, we demonstrate the versatility of ASGDRO, as it can easily be integrated with other algorithms. Moreover, our results show that ASGDRO not only improves performance in the context of subpopulation shift but also achieves performance gains in the presence of domain shift. For experimental details, we set the range of the robust step size $\gamma$ as \verb|lambda r: 10**r.uniform(-4, -2)| with $\gamma=0.001$ by default and the neighborhood size $\rho$ as \verb|lambda r: r.choice([0.05, 0.5, 1.0, 5.0])|. The other settings are the same as DPLCLIP \citep{zhang2021amortized}. Following common convention, we conduct 20 hyperparameter searches and reported the averages for three random seeds. We evaluated our model on the five datasets: VLCS \citep{fang2013unbiased}, PACS \citep{li2017deeper}, OfficeHome \citep{venkateswara2017deep},  TerraIncognita \citep{beery2018recognition} and DomainNet \citep{peng2019moment}. For the original ASGDRO experiments, we follow the experimental setup of \cite{wang2023sharpness}\footnote{https://github.com/Wang-pengfei/SAGM}.

\subsubsection*{Model selection: training-domain validation set}

\subsubsection*{VLCS}

\begin{center}
\adjustbox{max width=\linewidth}{%
\begin{tabular}{lccccc}
\toprule
\textbf{Algorithm}   & \textbf{C}           & \textbf{L}           & \textbf{S}           & \textbf{V}           & \textbf{Avg}         \\
\midrule
DPLCLIP              & 99.1 $\pm$ 0.5       & 61.1 $\pm$ 1.5       & 72.6 $\pm$ 2.6       & 83.1 $\pm$ 2.5       & 79.0                 \\
DPLCLIP GDRO         & 99.9 $\pm$ 0.0       & 61.3 $\pm$ 2.5       & 74.4 $\pm$ 1.1       & 83.4 $\pm$ 2.6       & 79.7                 \\
DPLCLIP ASGDRO       & 100.0 $\pm$ 0.0      & 62.7 $\pm$ 0.4       & 74.5 $\pm$ 1.4       & 85.7 $\pm$ 0.8       & 80.7                 \\
\bottomrule
\end{tabular}}
\end{center}

\subsubsection*{PACS}

\begin{center}
\adjustbox{max width=\linewidth}{%
\begin{tabular}{lccccc}
\toprule
\textbf{Algorithm}   & \textbf{A}           & \textbf{C}           & \textbf{P}           & \textbf{S}           & \textbf{Avg}         \\
\midrule
DPLCLIP              & 97.6 $\pm$ 0.2       & 98.3 $\pm$ 0.3       & 99.9 $\pm$ 0.0       & 90.5 $\pm$ 0.5       & 96.6                 \\
DPLCLIP GDRO         & 97.0 $\pm$ 0.7       & 98.2 $\pm$ 0.1       & 99.8 $\pm$ 0.1       & 88.6 $\pm$ 1.4       & 95.9                 \\
DPLCLIP ASGDRO       & 97.7 $\pm$ 0.1       & 98.7 $\pm$ 0.1       & 99.8 $\pm$ 0.0       & 91.0 $\pm$ 0.5       & 96.8                 \\
\bottomrule
\end{tabular}}
\end{center}

\subsubsection*{OfficeHome}
\vspace{-0.5em}
\begin{center}
\adjustbox{max width=\linewidth}{%
\begin{tabular}{lccccc}
\toprule
\textbf{Algorithm}   & \textbf{A}           & \textbf{C}           & \textbf{P}           & \textbf{R}           & \textbf{Avg}         \\
\midrule
DPLCLIP              & 80.6 $\pm$ 0.8       & 69.2 $\pm$ 0.2       & 90.1 $\pm$ 0.2       & 91.1 $\pm$ 0.0       & 82.7                 \\
DPLCLIP GDRO         & 82.3 $\pm$ 0.2       & 70.9 $\pm$ 0.1       & 90.0 $\pm$ 0.4       & 91.1 $\pm$ 0.1       & 83.6                 \\
DPLCLIP ASGDRO       & 82.1 $\pm$ 0.4       & 71.3 $\pm$ 0.8       & 90.3 $\pm$ 0.6       & 91.2 $\pm$ 0.3       & 83.7                 \\

\bottomrule
\end{tabular}}
\end{center}

\subsubsection*{TerraIncognita}

\begin{center}
\adjustbox{max width=\linewidth}{%
\begin{tabular}{lccccc}
\toprule
\textbf{Algorithm}   & \textbf{L100}        & \textbf{L38}         & \textbf{L43}         & \textbf{L46}         & \textbf{Avg}         \\
\midrule
DPLCLIP              & 47.1 $\pm$ 1.4       & 50.1 $\pm$ 1.2       & 41.6 $\pm$ 1.9       & 42.7 $\pm$ 0.7       & 45.4                 \\
DPLCLIP GDRO         & 49.1 $\pm$ 0.9       & 48.7 $\pm$ 2.6       & 46.3 $\pm$ 2.6       & 39.8 $\pm$ 1.4       & 46.0                 \\
DPLCLIP ASGDRO       & 52.8 $\pm$ 0.9       & 51.5 $\pm$ 2.1       & 49.2 $\pm$ 1.2       & 42.1 $\pm$ 0.9       & 48.9                 \\
\bottomrule
\end{tabular}}
\end{center}

\subsubsection*{DomainNet}

\begin{center}
\adjustbox{max width=\linewidth}{%
\begin{tabular}{lccccccc}
\toprule
\textbf{Algorithm}   & \textbf{clip}        & \textbf{info}        & \textbf{paint}       & \textbf{quick}       & \textbf{real}        & \textbf{sketch}      & \textbf{Avg}         \\
\midrule
DPLCLIP              & 70.9 $\pm$ 0.3       & 51.9 $\pm$ 0.3       & 66.6 $\pm$ 0.3       & 14.6 $\pm$ 0.5       & 84.3 $\pm$ 0.2       & 66.6 $\pm$ 0.1       & 59.1                 \\
DPLCLIP GDRO     & 71.8 $\pm$ 0.4       & 51.3 $\pm$ 0.4       & 67.0 $\pm$ 0.3       & 15.3 $\pm$ 0.2       & 84.4 $\pm$ 0.1       & 65.0 $\pm$ 0.9       & 59.1                 \\
DPLCLIP ASGDRO    & 71.5 $\pm$ 0.5       & 52.2 $\pm$ 0.4       & 67.5 $\pm$ 0.6       & 16.4 $\pm$ 0.2       & 84.7 $\pm$ 0.1       & 66.5 $\pm$ 0.2       & 59.8                 \\
\bottomrule
\end{tabular}}
\end{center}

\subsubsection*{Averages}

\begin{center}
\adjustbox{max width=\linewidth}{%
\begin{tabular}{lcccccc}
\toprule
\textbf{Algorithm}        & \textbf{VLCS}             & \textbf{PACS}             & \textbf{OfficeHome}       & \textbf{TerraIncognita}   & \textbf{DomainNet}        & \textbf{Avg}              \\
\midrule
DPLCLIP                   & 79.0 $\pm$ 0.7            & 96.6 $\pm$ 0.1            & 82.7 $\pm$ 0.2            & 45.4 $\pm$ 1.0            & 59.1 $\pm$ 0.1            & 72.6                      \\
DPLCLIP GDRO          & 79.7 $\pm$ 1.3            & 95.9 $\pm$ 0.4            & 83.6 $\pm$ 0.1            & 46.0 $\pm$ 1.0            & 59.1 $\pm$ 0.2            & 72.9                      \\
DPLCLIP ASGDRO         & 80.7 $\pm$ 0.3            & 96.8 $\pm$ 0.2            & 83.7 $\pm$ 0.5            & 48.9 $\pm$ 0.3            & 59.8 $\pm$ 0.2            & 74.0                      \\
\bottomrule
\end{tabular}}
\end{center}

% \vfill\eject

\subsubsection*{Model selection: test-domain validation set (Oracle)}

\subsubsection*{VLCS}

\begin{center}
\adjustbox{max width=\linewidth}{%
\begin{tabular}{lccccc}
\toprule
\textbf{Algorithm}   & \textbf{C}           & \textbf{L}           & \textbf{S}           & \textbf{V}           & \textbf{Avg}         \\
\midrule
DPLCLIP              & 99.8 $\pm$ 0.1       & 69.7 $\pm$ 0.6       & 72.4 $\pm$ 1.0       & 86.2 $\pm$ 0.5       & 82.0                 \\
DPLCLIP GDRO         & 99.9 $\pm$ 0.0       & 64.9 $\pm$ 1.1       & 79.1 $\pm$ 0.5       & 86.5 $\pm$ 0.2       & 82.6                 \\
DPLCLIP ASGDRO       & 99.8 $\pm$ 0.1       & 67.4 $\pm$ 0.9       & 78.1 $\pm$ 0.5       & 86.9 $\pm$ 0.1       & 83.1                 \\
\bottomrule
\end{tabular}}
\end{center}

\subsubsection*{PACS}

\begin{center}
\adjustbox{max width=\linewidth}{%
\begin{tabular}{lccccc}
\toprule
\textbf{Algorithm}   & \textbf{A}           & \textbf{C}           & \textbf{P}           & \textbf{S}           & \textbf{Avg}         \\
\midrule
DPLCLIP              & 97.6 $\pm$ 0.1       & 98.7 $\pm$ 0.3       & 99.8 $\pm$ 0.1       & 91.2 $\pm$ 0.3       & 96.8                 \\
DPLCLIP GDRO         & 97.4 $\pm$ 0.3       & 98.9 $\pm$ 0.2       & 99.8 $\pm$ 0.1       & 91.9 $\pm$ 0.3       & 97.0                 \\
DPLCLIP ASGDRO       & 97.7 $\pm$ 0.2       & 99.1 $\pm$ 0.0       & 99.9 $\pm$ 0.0       & 91.7 $\pm$ 0.3       & 97.1                 \\
\bottomrule
\end{tabular}}
\end{center}

\subsubsection*{OfficeHome}

\begin{center}
\adjustbox{max width=\linewidth}{%
\begin{tabular}{lccccc}
\toprule
\textbf{Algorithm}   & \textbf{A}           & \textbf{C}           & \textbf{P}           & \textbf{R}           & \textbf{Avg}         \\
\midrule
DPLCLIP              & 81.7 $\pm$ 0.2       & 70.9 $\pm$ 0.1       & 90.3 $\pm$ 0.3       & 90.7 $\pm$ 0.0       & 83.4                 \\
DPLCLIP GDRO         & 81.3 $\pm$ 0.8       & 70.6 $\pm$ 0.3       & 90.5 $\pm$ 0.1       & 90.9 $\pm$ 0.3       & 83.3                 \\
DPLCLIP ASGDRO       & 83.2 $\pm$ 0.4       & 71.7 $\pm$ 0.2       & 91.9 $\pm$ 0.1       & 91.3 $\pm$ 0.1       & 84.5                 \\
\bottomrule
\end{tabular}}
\end{center}

\subsubsection*{TerraIncognita}

\begin{center}
\adjustbox{max width=\linewidth}{%
\begin{tabular}{lccccc}
\toprule
\textbf{Algorithm}   & \textbf{L100}        & \textbf{L38}         & \textbf{L43}         & \textbf{L46}         & \textbf{Avg}         \\
\midrule
DPLCLIP              & 55.9 $\pm$ 2.3       & 58.5 $\pm$ 0.3       & 48.2 $\pm$ 0.5       & 40.9 $\pm$ 3.0       & 50.9                 \\
DPLCLIP GDRO         & 57.9 $\pm$ 1.0       & 55.3 $\pm$ 1.5       & 49.6 $\pm$ 2.0       & 41.8 $\pm$ 1.4       & 51.2                \\
DPLCLIP ASGDRO       & 56.2 $\pm$ 0.8       & 54.1 $\pm$ 0.3       & 50.7 $\pm$ 0.7       & 42.1 $\pm$ 0.5       & 50.8                 \\
\bottomrule
\end{tabular}}
\end{center}

\subsubsection*{DomainNet}

\begin{center}
\adjustbox{max width=\linewidth}{%
\begin{tabular}{lccccccc}
\toprule
\textbf{Algorithm}   & \textbf{clip}        & \textbf{info}        & \textbf{paint}       & \textbf{quick}       & \textbf{real}        & \textbf{sketch}      & \textbf{Avg}         \\
\midrule
DPLCLIP              & 72.0 $\pm$ 0.5       & 52.1 $\pm$ 0.3       & 67.3 $\pm$ 0.2       & 16.6 $\pm$ 0.2       & 84.4 $\pm$ 0.2       & 66.8 $\pm$ 0.1       & 59.9                 \\
DPLCLIP GDRO         & 72.0 $\pm$ 0.2       & 51.7 $\pm$ 0.1       & 67.2 $\pm$ 0.4       & 16.7 $\pm$ 0.2       & 84.5 $\pm$ 0.0       & 66.3 $\pm$ 0.1       & 59.7                 \\
DPLCLIP ASGDRO       & 71.5 $\pm$ 0.5       & 52.8 $\pm$ 0.3       & 68.1 $\pm$ 0.3       & 16.5 $\pm$ 0.2       & 84.9 $\pm$ 0.0       & 67.0 $\pm$ 0.1       & 60.2                 \\
\bottomrule
\end{tabular}}
\end{center}

\subsubsection*{Averages}

\begin{center}
\adjustbox{max width=\linewidth}{%
\begin{tabular}{lcccccc}
\toprule
\textbf{Algorithm}        & \textbf{VLCS}             & \textbf{PACS}             & \textbf{OfficeHome}       & \textbf{TerraIncognita}   & \textbf{DomainNet}        & \textbf{Avg}              \\
\midrule
DPLCLIP                   & 82.0 $\pm$ 0.3            & 96.8 $\pm$ 0.1            & 83.4 $\pm$ 0.1            & 50.9 $\pm$ 0.6            & 59.9 $\pm$ 0.2            & 74.6                      \\
DPLCLIP GDRO              & 82.6 $\pm$ 0.2            & 97.0 $\pm$ 0.2            & 83.3 $\pm$ 0.2            & 51.2 $\pm$ 1.0            & 59.7 $\pm$ 0.0            & 74.8                      \\
DPLCLIP ASGDRO            & 83.1 $\pm$ 0.2            & 97.1 $\pm$ 0.1            & 84.5 $\pm$ 0.1            & 50.8 $\pm$ 0.3            & 60.2 $\pm$ 0.1            & 75.1                      \\

\bottomrule
\end{tabular}}
\end{center}

\subsection{Grad-CAM Analysis} \label{append:gradcam}

\begin{figure}[h!]
    \centering
        \includegraphics[height=0.85\textheight,keepaspectratio]{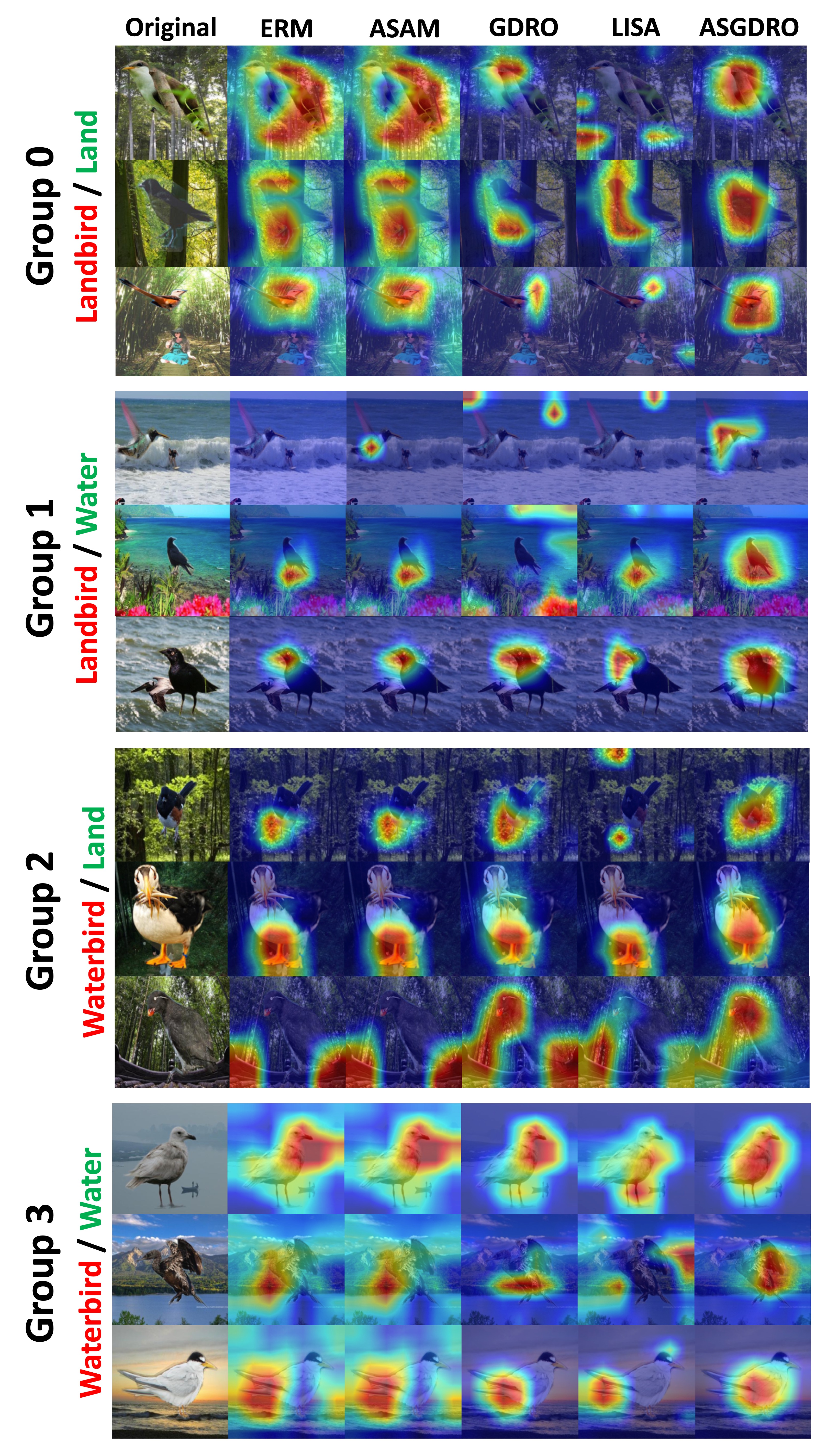}
    \caption{\textbf{Grad-CAM results on the Waterbirds Dataset.} 
The words highlighted in red represent invariant features: Landbird and Waterbird. On the contrary, the words highlighted in green represent spurious features: Land and Water background. In the Training Set, Group 1 and Group 2 are minority groups with significantly fewer data samples compared to other groups.}
    \label{fig:cub_gradcam}
\end{figure}

\begin{figure}[h!]
    \centering
        \includegraphics[height=0.85\textheight,keepaspectratio]{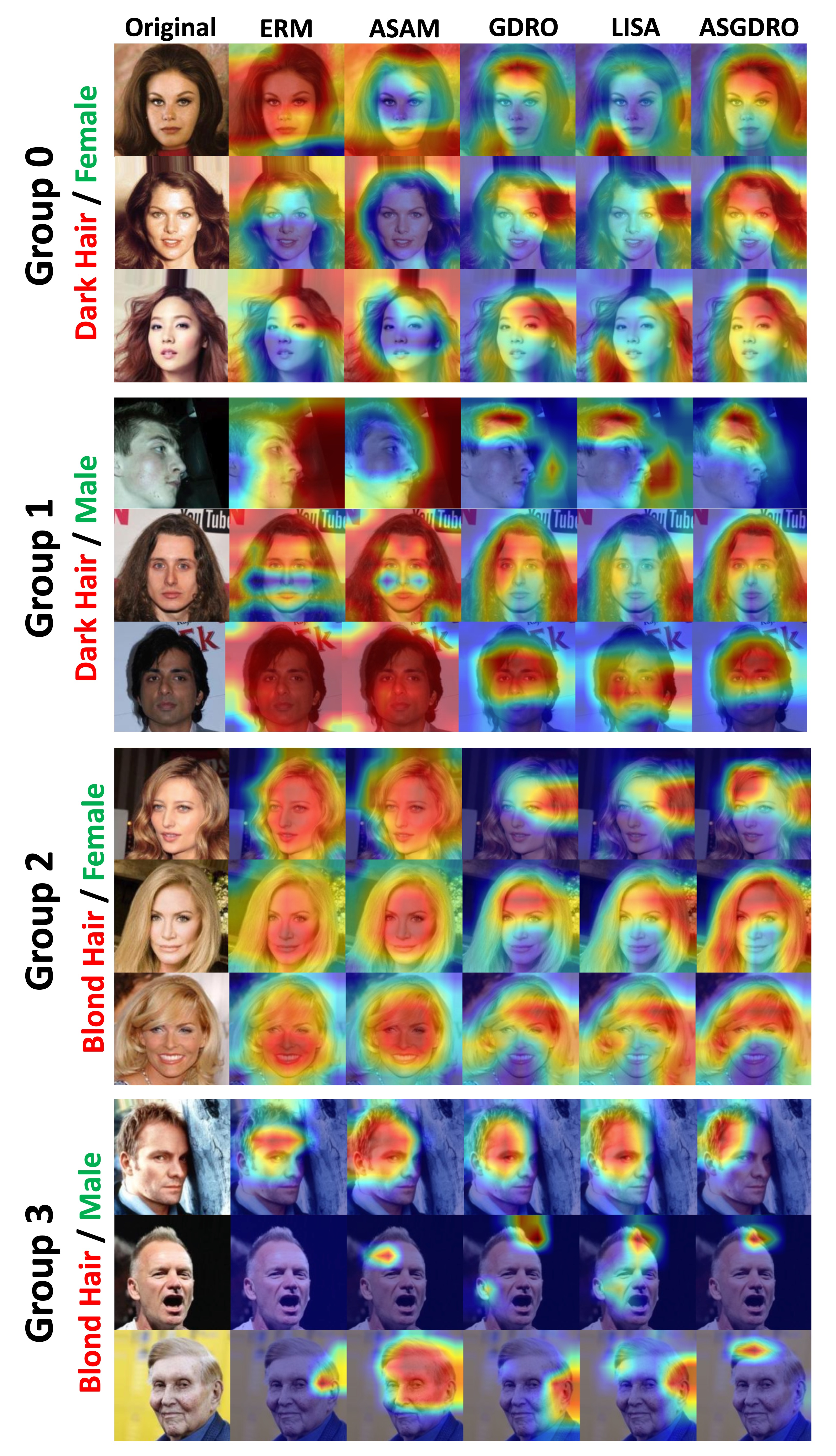}
    \caption{\textbf{Grad-CAM results on the CelebA Dataset.} The features highlighted in red represent invariant words: Dark Hair and Blond Hair. On the contrary, the words highlighted in green represent spurious features: Female and Male. In the Training Set, Group 3 is a minority group with significantly fewer data samples compared to other groups.}
    \label{fig:celeba_gradcam}
    \vspace{-1.0em}
\end{figure}

In this section, we present additional Grad-CAM \citep{Selvaraju@gradcam} results on the Waterbirds and CelebA datasets. In Figure \ref{fig:cub_gradcam} and \ref{fig:celeba_gradcam}, the red-colored-name features represent invariant features in the respective task, while the green-colored-name features represent spurious features. In the Grad-CAM images, the pixels that each model focuses on to predict the ground-truth label are highlighted closer to the red color in the image.

ERM \citep{vapnik99} and ASAM \citep{kwon2021asam} are regularization-free algorithms that do not specifically encourage models to focus on invariant features, and this is reflected in the Grad-CAM results. Specifically, when observing Group 0 and Group 3 of Waterbirds, which can strongly form the correlation between class and spurious, as well as Group 0, 1, and 2 of CelebA, in most cases, the results show a strong focus on both spurious and invariant features simultaneously or solely on spurious features. For some images, particularly between CelebA dataset's Group 0 and 1 where there are no minority groups within a class, there is some degree of focus on invariant features. However, these images still contain a significant amount of unnecessary pixels such as the background. Conversely, in minority groups such as Group 1 and 2 in Waterbirds or Group 3 in CelebA, there is a predominant focus on invariant features to predict the ground-truth label. However, this focus is limited to only a subset of the overall invariant features and still include some spurious features.

In algorithms specifically designed to learn invariant features like GDRO \citep{sagawa2019distributionally}, LISA \citep{yao2022improving}, and ASGDRO (Ours), the Grad-CAM results exhibit different patterns compared to ERM and ASAM. In the most of results for the three algorithms, the models demonstrate a reasonable focus on invariant features. Compared with ERM and ASAM, there are significant reductions in the extent to which they focus on spurious features. However, GDRO and LISA still concentrate only on a part of invariant features. Additionally, in some cases, they may exhibit a greater focus on spurious features than on the subset of invariant features. It is also frequently observed that they still heavily include spurious features or solely focus on spurious features when dealing with majority groups such as Group 1 and 3 in Waterbirds or Group 0, 1, and 2 in CelebA. As in the results of Group 1, and 2 in Waterbirds or Group 3 in CelebA, we observe that the models' low ability to fully concentrate on invariant features is affected by the performance of models that still exhibit a focus on spurious features. This observation highlights the impact of the models' performance on their ability to completely focus on invariant features. 

In contrast to other baselines, ASGDRO demonstrates a stronger focus on invariant features. As a result, Grad-CAM analysis shows that ASGDRO has relatively larger regions of focus on invariant features compared to other baselines. Simultaneously, it successfully eliminates spurious features while accurately predicting the ground-truth label. Therefore, these results demonstrate that ASGDRO has a higher capacity for capturing sufficiently diverse invariant features, and this characteristic is reflected in its performance. That is, ASGDRO promotes that the model performs SIL.

\newpage

\newpage

\newpage

\subsection{Hessian Analysis for Waterbirds Dataset} \label{append:hess}

\begin{table}[h]
\centering
\begin{adjustbox}{width=0.8\linewidth}
\begin{tabular}{cccccccc}
\toprule
& \multicolumn{3}{c}{The Largest Eigenvalue} && \multicolumn{3}{c}{The Second Largest Eigenvalue} \\ \cmidrule(lr){2-4} \cmidrule(lr){6-8}
Method & Majority      & Minority      & Total   &   & Majority         & Minority        & Total        \\ \midrule
ERM    & 990           & 4894          & 2265    &   & 166              & 511             & 709          \\ \midrule
ASAM   & 972           & 5475          & 2624    &  & 178              & 524             & 647          \\ \midrule
GDRO   & 131           & 447           & 353     & & 118              & 346             & 129          \\ \midrule
ASGDRO & \textbf{107}           & \textbf{342}           & \textbf{279}     & & \textbf{98}               & \textbf{274}             & \textbf{105}          \\ \bottomrule
\end{tabular}
\end{adjustbox}
\vspace{0.5em}
\caption{\small \textbf{Hessian Analysis on Waterbirds.} ASGDRO finds the common flat minima for both majority and minority groups.} \label{table:hess_waterbirds}
\end{table}
ERM and ASAM have significantly sharper minima for the minority group compared to GDRO and ASGDRO due to the spurious correlation, although ASAM is designed to find flat minima. Compared to GDRO and other baselines, ASGDRO achieves the lowest eigenvalue in the first and second maximum eigenvalues for every group.

\end{document}